\documentclass[10pt,twocolumn,letterpaper]{article}

\usepackage{multirow} % For \multirow command
\usepackage[pagenumbers]{cvpr} % To force page numbers, e.g. for an arXiv version
\usepackage{float}
\usepackage{afterpage}
\usepackage[accsupp]{axessibility}  % Improves PDF readability for those with disabilities.
\definecolor{cvprblue}{rgb}{0.21,0.49,0.74}
\usepackage[pagebackref,breaklinks,colorlinks,allcolors=cvprblue]{hyperref}
\usepackage{bbm}
\usepackage{algorithm}
\usepackage{algorithmic}

\def\paperID{8639} % *** Enter the Paper ID here
\def\confName{CVPR}
\def\confYear{2025}

\makeatletter
  \def\title@font{\Large\bfseries}
  \let\ltx@maketitle\@maketitle
  \def\@maketitle{\bgroup%
    \let\ltx@title\@title%
    \def\@title{\resizebox{\textwidth}{!}{%
      \mbox{\title@font\ltx@title}%
    }}%
    \ltx@maketitle%
  \egroup}
\makeatother

\makeatletter
\let\oldmaketitlesupplementary\maketitlesupplementary
\renewcommand{\maketitlesupplementary}{
  \begingroup 
  \let\oldthetitle\thetitle
  \renewcommand{\thetitle}{%
    \resizebox{\textwidth}{!}{%
      \mbox{\title@font\oldthetitle}%
    }
  }
  \oldmaketitlesupplementary
  \endgroup
}
\makeatother

\title{H2ST: Hierarchical Two-Sample Tests for Continual Out-of-Distribution Detection}

\author{Yuhang Liu$^{1}$ \hspace{1cm} Wenjie Zhao$^{2}$ \hspace{1cm} Yunhui Guo$^{2}$\\
$^{1}$University of Electronic Science and Technology of China \hspace{1cm} $^{2}$ University of Texas at Dallas\\
$^{1}${\tt\small yuhang.liu.ce@gmail.com}, $^{2}${\tt\small \{wxz220013,  yunhui.guo\}@utdallas.edu}
}
\begin{document}
\maketitle
\begin{abstract}
Task Incremental Learning (TIL) is a specialized form of Continual Learning (CL) in which a model incrementally learns from non-stationary data streams. Existing TIL methodologies operate under the closed-world assumption, presuming that incoming data remains in-distribution (ID). However, in an open-world setting, incoming samples may originate from out-of-distribution (OOD) sources, with their task identities inherently unknown. Continually detecting OOD samples presents several challenges for current OOD detection methods: reliance on model outputs leads to excessive dependence on model performance, selecting suitable thresholds is difficult, hindering real-world deployment, and binary ID/OOD classification fails to provide task-level identification. To address these issues, we propose a novel continual OOD detection method called the Hierarchical Two-sample Tests (H2ST). H2ST eliminates the need for threshold selection through hypothesis testing and utilizes feature maps to better exploit model capabilities without excessive dependence on model performance. The proposed hierarchical architecture enables task-level detection with superior performance and lower overhead compared to non-hierarchical classifier two-sample tests. Extensive experiments and analysis validate the effectiveness of H2ST in open-world TIL scenarios and its superiority to the existing methods. Code is available at \href{https://github.com/YuhangLiuu/H2ST}{https://github.com/YuhangLiuu/H2ST}.
\end{abstract}
    
\section{Introduction}
\label{sec:intro}
Continual learning (CL) \cite{de2021cl, chen2018lifelong, wang2024comprehensive, aljundi2017expert}, also known as lifelong learning, is built on the idea of learning continuously about the external world to enable the incremental development of ever more complex knowledge. Task incremental learning (TIL) \cite{hossain2022rethinkingtil, de2021cl, type_til} is a specific form of CL.
\par
Existing TIL methods \cite{bang2021rainbow, hyder2022incremental, jiang2023neural, gem} operate under the closed-world assumption, where test data is assumed to be drawn i.i.d. from the same distribution as the training data, known as in-distribution (ID). However, when deployed in an open-world scenario \cite{open_world}, models may encounter test samples from out-of-distribution (OOD) sources, which require careful handling. This is particularly critical in safety-sensitive applications such as healthcare and autonomous driving, where mishandling OOD samples can lead to severe consequences. Moreover, task identity (task-id) is not explicitly provided and needs to be predicted in open-world.
\par
To detect OOD samples, existing OOD detection methods can be considered. These include utilizing the maximum softmax probability (MSP) as the indicator score \cite{msp}, using temperature scaling and input perturbation to amplify the separability \cite{odin}, and using energy score \cite{energy} as potential indicators.
However, these traditional OOD detection methods are not specifically tailored for TIL; they assume a static set of ID classes which starkly contrasts with the dynamic nature of TIL, wherein an OOD class may transition into an ID class at subsequent stages. In particular, their limitations are as follows. (1) The above methods only produce an indicator score, necessitating a threshold to distinguish between ID and OOD samples. 
%Moreover, empirical thresholds are often suboptimal.
(2) The effectiveness of OOD detection depends on the performance of the model \cite{kim2022theoretical}. In particular, existing methods largely hinge on the derived probability or logit outputs of the model \cite{msp, energy, maxlogit}. If the model performs poorly on ID samples, its ability to detect OOD samples will also be compromised. (3) These OOD detection methods only achieve binary classification of ID and OOD, without discerning among ID tasks. For instance, all ID tasks ideally have a numerically higher MSP score than OOD, but there is no discernible separation between different ID tasks. However in open-world TIL, without task-id, the model will not be able to perform inference.
\par
The two-sample test is a statistical hypothesis test designed to determine whether two samples are drawn from the same distribution without requiring thresholds. Classifier two-sample tests (C2ST) \cite{lopez2016revisitingc2st} train a binary classifier, using its accuracy as the test statistic. The value of the test statistic is compared to a critical value also obtained by statistical methods to ascertain whether the null hypothesis can be rejected. Nevertheless, C2ST does not apply to continual OOD detection, due to the 
following limitations. Raw samples lead to high-dimensional inputs, raising costs, and missing high-level semantics. A single binary classifier lacks task-id identification, while using a specific classifier per task complicates the classification and increases overhead as the number of tasks grows.
\par To address the limitations of traditional OOD detection methods and C2ST, we propose Hierarchical Two-Sample Tests (H2ST) for continual OOD detection. H2ST utilizes feature maps rather than relying on samples or model outputs to perform two-sample tests, eliminating the need for thresholds and mitigating over-dependence on the model. Its hierarchical architecture facilitates fine-grained task-ID prediction while enhancing detection performance with reduced computational overhead. Extensive experiments and analysis validate the superiority of H2ST in open-world TIL scenarios.
\par 
Overall, the contributions of this paper are as follows:
\begin{itemize}
    \item We are the first to introduce the open-world TIL scenario, where TIL models operate in an open-world setting. This allows models to detect OOD samples, predict ID samples in unfamiliar contexts, and incrementally expand their knowledge base.
    \item We propose Hierarchical Two-sample Tests (H2ST) for continual OOD detection, enabling a threshold-free, fine-grained assessment of sample outlier status while minimizing overhead.
    \item Through extensive experiments, we demonstrate H2ST's compatibility with replay-based TIL methods and superior performance over existing OOD detection methods.
\end{itemize}
\section{Related Work}
\label{sec:related}
\subsection{Continual Learning (CL)}
CL is a research area focused on developing models that can learn and adapt to new tasks over time without forgetting previously acquired knowledge. The key challenge in CL is catastrophic forgetting (CF) \cite{1989CF, evron2022cf}, where learning new tasks degrades performance on previously learned tasks. To handle the CF issue, the proposed CL methods can be categorized into three types: (1) Replay-based methods store a few old training samples within a memory buffer \cite{gem,guo2020improved}. For example, GEM \cite{gem} establishes individual constraints based on the trained samples to ensure a non-increase in losses. FeTrIL \cite{petit2023fetril} utilizes a fixed feature extractor derived from the initial task and subsequently replays the generated features. (2) Regularization-based methods \cite{kirkpatrick2017overcoming, ritter2018online,yu2024evolve} add explicit regularization terms to balance the old and new tasks. For example, the importance of each parameter is measured by evaluating its contribution to the total loss variation \cite{zenke2017continualSI} and the sensitivity of predictive results to parameter changes \cite{aljundi2018memoryAS}. (3) Architecture-based methods construct task-specific parameters with a carefully designed architecture. HAT \cite{serra2018overcomingHAT} and WSN \cite{kang2022forgetWSN} explicitly optimize a binary mask to select specific neurons or parameters for each task, with the masked regions of previous tasks frozen. These methods are all based on the closed-world assumption, so if there is an OOD sample during testing, the CL model will produce a wrong prediction rather than reporting it as OOD. In this paper, we extend TIL to open-world, where we face dual uncertainties: whether new samples are ID or OOD, and their corresponding task-ids. To address these challenges, we incorporate OOD detection into TIL to handle the OOD issue and predict task-id at the same time.

\subsection{Out-of-Distribution (OOD) Detection}
OOD detection \cite{yang2024generalizedood} plays a pivotal role in ensuring the reliability and safety of machine learning systems. Recent focus has shifted to post-hoc OOD scoring methods, easy to use without requiring changes to the training procedure. The original baseline uses the maximum softmax probability (MSP) as score \cite{msp}. Expanding on this, ODIN \cite{odin} combines input preprocessing and temperature scaling to enhance the differentiation. The energy score \cite{energy} utilizes an energy-based model to reduce prediction bias, benefiting from a theoretical interpretation grounded in the likelihood perspective \cite{morteza2022provable}. ReAct \cite{sun2021react}, NMD \cite{dong2022nmd}, and LHACT \cite{yuan2023lhact} refine the output features through activation rectification, while DICE \cite{sun2022dice} leverages sparsification to achieve stronger separability between ID and OOD samples. GradNorm \cite{huang2021gradnorm} and FeatureNorm \cite{feature_norm} respectively utilize the vector norm of gradients and feature maps. However, choosing suitable thresholds for these scoring methods to distinguish ID and OOD samples presents difficulties.
Additionally, these methods predominantly focus on static scenarios instead of dynamic ones like CL. Additionally, the CF in CL can degrade the performance of OOD detection.  Moreover, in the open-world TIL scenario, it is necessary not only to distinguish between ID and OOD but also to determine the task-ids of predicted ID samples, which these methods fail to provide. To address these limitations, we propose a task-specific hierarchical detection method at the feature level.
 
\subsection{Continual Out-of-Distribution Detection}
The study of OOD detection in the CL scenario has garnered increasing attention from the research community, with the majority of studies adopting the class incremental learning (CIL) paradigm. The pioneering work \cite{aljundi2022continualnovelty} introduces the problem of continual novelty detection by deploying novelty detection methods in a CL scenario. OpenCIL \cite{miao2024opencil} evaluates the capability of different OOD methods in various CIL models; however, CIL and OOD remain largely decoupled, with the performance of OOD detection exerting no influence on CIL. In \cite{he2022out}, a method is proposed to correct output bias and enhance the confidence difference between ID and OOD data in unsupervised continual learning. CEDL \cite{aguilar2023cedl} incorporates a deep evidential learning approach within a CL framework to perform OOD detection based on uncertainty estimation.
In \cite{kim2024OCL}, the authors decompose CIL into two components: within-task prediction (corresponds to TIL), and task-id prediction (associated with OOD detection). 
%We show this by decomposing CIL into two sub-problems: within-task prediction (WP) and task-id prediction (TP), and proving that TP is correlated with closed-world OOD detection. 
%WP means that the prediction for a test instance is only made within the classes of the task to which the test instance belongs, which is basically the TIL problem.
They further provide theoretical proof that both are essential for achieving optimal CIL performance. 
%good WP and good closed-world OOD detection are necessary and sufficient conditions for good CIL
Based on these theoretical results, MORE \cite{kim2024OCL, kim2022multi} treats the saved samples from learned tasks as OOD samples in learning new tasks, implementing closed-world OOD detection. 
%the OOD detection method used in this section is a closed-world method as it uses the saved samples from previous tasks as OOD samples in learning each new task.
%The OOD detection method used there is a closed-world OOD detection method as it treats the replay data from previous tasks as the OOD data in the model building
Their core focus is on using OOD detection as a tool to leverage TIL for achieving CIL, rather than general open-world OOD detection. The key distinction between these CIL approaches and open-world TIL is that the tasks can be independent and even have overlapping labels in open-world TIL, ensuring our setting can be broadly applicable. For instance, consider a medical system that incrementally learns lung X-ray classification (normal, pneumonia, and abnormal) and liver CT analysis (normal, cirrhosis, and abnormal). The label overlap (\eg, "normal" and "abnormal" in both tasks) creates an inherent incompatibility with CIL that assumes disjoint label spaces. Therefore, the proposed setting is more general than open-world CIL.
\section{Background}
\label{sec:bg}

\subsection{Task Incremental Learning}
\label{sub_sec:til}
Task incremental learning (TIL) can be formulated as a sequence of $N$ training tasks, each with a corresponding training set $ \{D_{1}, D_{2}, \ldots, D_{N} \}$, where $D_{n}=\{(x_{i}, t_{i}, y_{i})\}_{i=1}^{P_{n}}$ represents the $n$-th training set with $P_{n}$ training instances. Here, $x_{i} \in \mathcal{X}_{n}$ is a sample, $t_{i}$ is the task-id, and $y_{i} \in \mathcal{Y}_{n}$ is the corresponding label. Unlike CIL requiring disjoint label spaces, TIL allows for more flexibility in task definitions, including disjoint or overlapping label spaces. Our goal is to learn a model $f_\theta$ parameterized by weights $\theta$ by minimizing the classification loss $\arg \min_{\theta} \mathcal{L}\left(f_\theta\left(x, t\right), y\right)$, where $\mathcal{L}$ denotes the loss function. In addition, we use replay-based TIL methods, storing $M_{n}$ samples from the $n$-th task in a memory buffer $B_{n}$. As stated earlier, we consider the more practical open-world TIL setting. Unlike closed-world TIL, task-id is not explicitly given and needs to be predicted. 

\subsection{Continual OOD Detection}
\label{sub_sec:cood}
In continual OOD Detection, in each step, the test samples contain both ID and OOD samples, so for each test sample. Therefore, an OOD detector $d$ is needed to determine whether the input sample $x$ is OOD or not. A prevalent strategy is to calculate the score $\varphi \left(x; f_\theta \right)$ of the sample $x$ and then make a judgment based on a threshold, 
\begin{align}
d\left(x ; f_\theta\right) & = \left\{\begin{array}{ll}
\text{\textit{ID}} &\text{if } \varphi\left(x ; f_\theta\right) \leq \gamma \\
\text{\textit{OOD}} &\text{if } \varphi\left(x ; f_\theta\right)>\gamma ,\nonumber
\end{array}\right.
\end{align}
where $\varphi$ can be any OOD score function and $\gamma$ is the threshold. 
For samples predicted as ID, their labels will be further predicted using the TIL model $f_\theta$. For samples predicted as OOD, we directly classify them as OOD class. The  formulation of continual OOD is as follows:
\begin{align}
h\left(x; f_\theta\right) & = \left\{\begin{array}{ll}
\underset{i}{\arg \max } f_\theta(x, t) & \text { if } d\left(x ; f_\theta\right)  = \text{\textit{ID}} \\
\text{\textit{OOD}} & \text { if } d\left(x ; f_\theta\right)  =\text{\textit{OOD}},
\end{array}\right. \nonumber
\end{align}
where $\underset{i}{\arg \max } f_\theta(x, t)$ indicates the classification result, $d$ is the OOD detector, and $t$ is the task-id obtained in the OOD detection stage. After the OOD samples are labeled, they can be used as new training samples, and the model $f_\theta$ can start the next round of incremental training.

Existing OOD detection methods face several limitations, as they require threshold selection, depend on model performance, and cannot discern tasks within ID samples. To address these issues, we propose a specialized OOD detection method called Hierarchical Two-Sample Test, inspired by classic two-sample tests in statistical hypothesis testing.
\subsection{Two-Sample Tests}
\label{sub_sec:2st}
The goal of two-sample tests is to assess whether two samples are drawn from the same distribution.  Based on this, the idea of classifier two-sample tests (C2ST) 
\cite{lopez2016revisitingc2st} is to train a binary classifier that distinguishes between source and target samples. A test statistic is derived from the classifier accuracy and a univariate two-sample test determines the cutoff value for rejecting the null hypothesis \cite{friedman2003multivariate}. Recently, a sequential covariate shift detection method \cite{jang2022sequential} is proposed, using C2ST to check whether the current test samples differ in distribution compared to the training samples. At each detection time $\tau$, there will be a source sample $x_{\tau}$ drawn from a fixed set of source samples $\mathcal{S}$ and a new observed target sample $x_{\tau}^{\prime} \in \mathcal{T}_{\tau}$. In particular, their method consists of the following three steps.
\par \noindent
\textbf{Step 1. Source-target Prediction.} Source-target labels on the current samples $x_{\tau}$ and $x_{\tau}^{\prime}$  are predicted using the current source-target classifier $g$. Denote the predictions as $\hat{y}_{\tau}=g\left(x_{\tau}\right)$ and $\hat{y}_{\tau}^{\prime}=g\left(x_{\tau}^{\prime}\right)$.
\par \noindent
\textbf{Step 2. Calibrated Covariate Shift Detection.} Identify as covariate shift, if $1/2 \notin \Theta_{\mathrm{CP}}\left(2 w \hat{\mu}_{w, \tau}, 2 w ; \alpha\right) $, and no shift otherwise. The formulation of detector $d$ is as follows:
\begin{equation}
d\left(x_{\tau}, x_{\tau}^{\prime} ; w, \alpha\right)=\mathbbm{1}\left(\frac{1}{2} \notin \Theta_{\mathrm{CP}}\left(2 w \cdot \hat{\mu}_{w, \tau}, 2 w ; \alpha\right)\right),    \nonumber 
\end{equation}
where $w \in \mathbbm{N}$ denotes the window size, $\alpha$ denotes the significance level, $\Theta_{\mathrm{CP}}$ denotes the Clopper-Pearson (CP) interval, and 
\begin{align}
\hat{\mu}_{w, \tau} = \frac{1}{2 w} \sum_{i = \tau-w+1}^{\tau}\left(\mathbbm{1}\left(\hat{y}_{i} = y_{i}\right)+\mathbbm{1}\left(\hat{y}_{i}^{\prime} = y_{i}^{\prime}\right)\right) \nonumber
\end{align} denotes the unbiased empirical
estimate of the accuracy of $\hat{g}$ at distinguishing whether an sample $x$ is from distribution $\mathcal{S}$ or $\mathcal{T}_{\tau}$.
The CP interval can be calculated using the following equivalent formula: 
\begin{align}  
\Theta_{\mathrm{CP}}(\hat{s}, n ; \alpha)  = &\left[Q\left(\frac{\alpha}{2} ; \hat{s}, n-\hat{s}+1\right),\right. \nonumber \\
&\left.\quad Q\left(1-\frac{\alpha}{2} ; \hat{s}+1, n-\hat{s}\right)\right], \nonumber
\end{align}
where $Q(p, a, b)$ is the $p$-th quantile of a Beta distribution with parameters $a, b$ \cite{hartley1951chart, brown2001interval}.
\par \noindent
\textbf{Step 3. Online Source-target Classifier Update.} The binary source-target classifier $g$ will be updated using the source and target samples, $i.e.$, $\left(x_{\tau}, 0\right)$  and  $\left(x_{\tau}^{\prime}, 1\right)$
\par 
However, this method suffers from critical limitations when applied to OOD detection in open-world TIL. First, it necessitates additional storage resources for retaining source samples, and processing raw data directly results in high-dimensional inputs that not only increase computational costs but also miss high-level semantic representations. Furthermore, this method is not designed for continual learning and has a fixed covariate shift detection mechanism with only a single binary classifier. With the continuous expansion of ID tasks in TIL frameworks, it presents a critical dilemma: A single unified classifier architecture, although computationally efficient, fundamentally lacks task-id prediction capability due to its binary classification nature, we refer to this approach as single-C2ST. Conversely, maintaining a specific classifier for each task effectively preserves task discrimination at the expense of increased resource overhead with accumulating tasks; this is what we subsequently refer to as C2ST.
%\section{Methodology}
\label{sec:method}
\section{Proposed Approach}
\label{sub_sec:hierarchical c2st}
\begin{figure}[t]
  \centering
   \includegraphics[width=\linewidth]{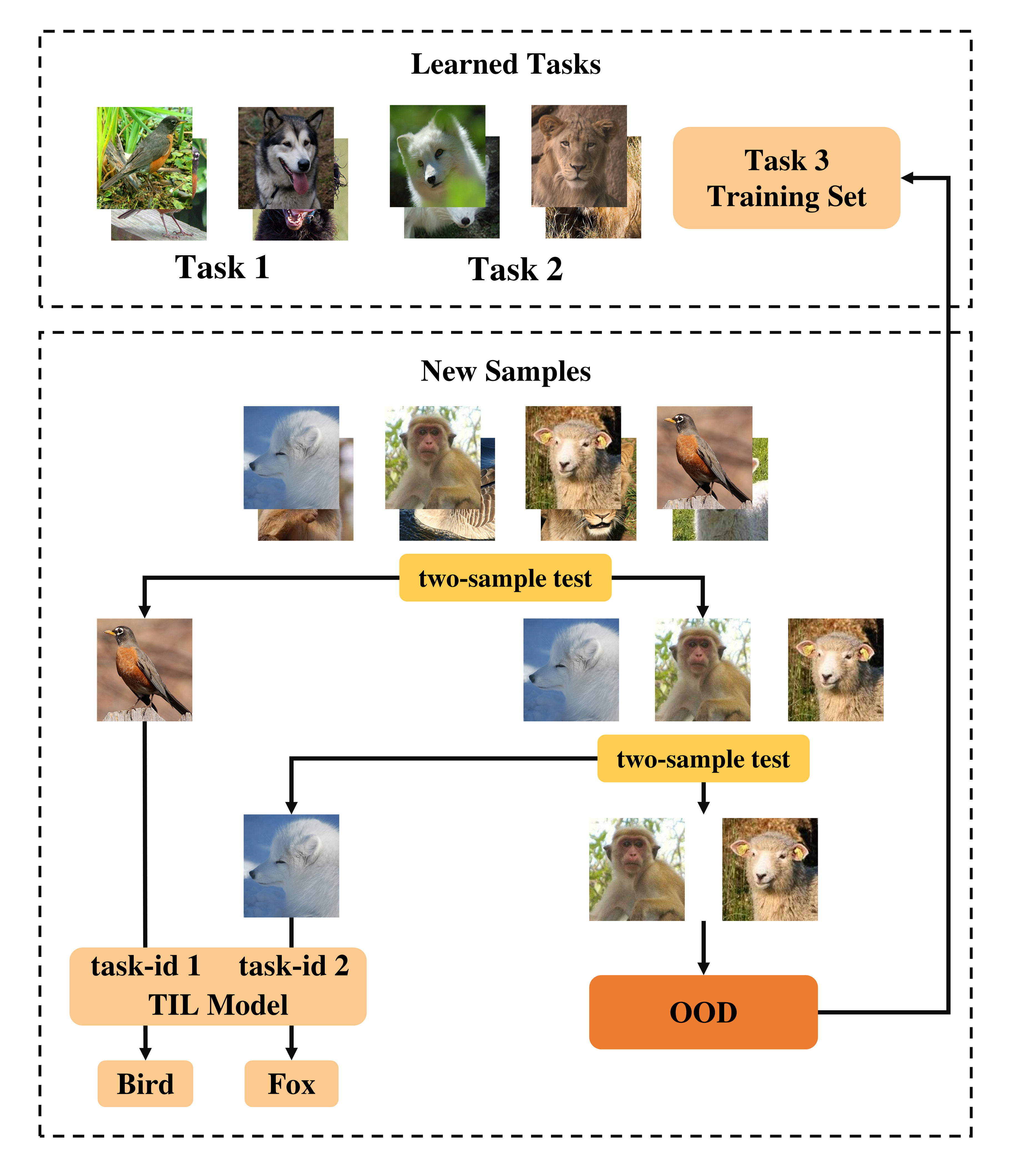}
   \caption{Illustration of H2ST for continual OOD detection. The TIL model learns incrementally and then proceeds to the testing phase. The new samples will traverse the hierarchical architecture with early-exit upon ID identification or completion of all layers. Samples predicted as ID will be further inferred, and the OOD will be used as the training samples for the new task.}
   \label{fig:method}
\end{figure}
In this paper, we propose a novel continual OOD detection method called the Hierarchical Two-Sample Tests (H2ST). H2ST is composed of $T$ task-specific two-sample test layers that are cascaded together, with each layer equipped with a source-target classifier initialized via Xavier initialization \cite{glorot2010understanding}. The architecture contains $T$ classifiers, $ \left\{g_{1}, g_{2}, \ldots, g_{T}\right\}$, where $T$ is the number of learned task. Each layer consists of three sequential steps: (1) feature-level source-target prediction, (2) online source-target classifier update, and (3) calibrated detection.  The following and Algorithm \ref{alg:h2st} include details. At each detection time $\tau$, there will be new target sample $x_{\tau}^{\prime}$ to be detected. After obtaining the feature maps of these current new samples and samples randomly drawn from memory buffers $ \left\{B_{1}, B_{2}, \ldots, B_{T}\right\}$, we perform hierarchical OOD detection by using the corresponding source-target classifiers in a progressive strategy. The sample $x_{\tau}^{\prime}$ traverses the hierarchical architecture, starting from the first layer, with early-exit upon confident ID identification or completion of all $T$ layers assessments. The details of the three steps for the $j$-th layer are as follows.
\par \noindent
\textbf{Step 1. Feature-level Source-Target Prediction.} We first randomly draw  $x_{\tau, j}$ form the memory buffer $B_{j}$ as the source sample. To avoid imbalance, the number of samples drawn from the memory is matched to the number of new samples. Then we obtain the feature map of the source sample and target sample $x_{\tau}^{\prime}$ using the TIL model $f_\theta$, denoted as $\psi\left(x_{\tau, j}\right)$ and $\psi\left(x_{\tau}^{\prime}\right)$. Then source-target labels on $\psi\left(x_{\tau, j}\right)$ and $\psi\left(x_{\tau}^{\prime}\right)$ are predicted using the corresponding source-target classifier $g_{j}$, denoted as $\hat{y}_{\tau, j}$ and $\hat{y}_{\tau, j}^{\prime}$. Unlike C2ST, which directly uses raw samples, H2ST instead employs feature representations as input to the classifier $g_{j}$, $i.e.$, $\hat{y}_{\tau, j}=g_{j}\left(\psi\left(x_{\tau, j}\right)\right)$ and $\hat{y}_{\tau, j}^{\prime}=g_{j}\left(\psi\left(x_{\tau}^{\prime}\right)\right)$, and we do not use the output of the TIL model $f_\theta$ as most OOD detection methods do. By doing so, H2ST reduces over-dependence on the model performance and optimizes the utilization of high-level feature representations.
\par \noindent
\textbf{Step 2. Online Source-target Classifier Update.} After getting the source-target predictions, the source-target classifier $g_{j}$ will be updated by treating source samples as 0 and target samples as 1, $i.e.$, $\left(x_{\tau, j}, 0\right)$  and  $\left(x_{\tau}^{\prime}, 1\right)$.
\par \noindent
\textbf{Step 3. Calibrated Detection.}  For the TIL model, there are totally $T$ ID tasks, but for the $j$-th two-sample test layer, only the $j$-th task is considered ID. Consequently, the detector $d_{j}$ in the $j$-th layer can only determine whether the new sample $x_{\tau}^{\prime}$ is OOD for task $j$. The formulation of $d_{j}$ is thus as follows:
\begin{equation}
d_{j}\left(x_{\tau, j}, x_{\tau}^{\prime} ; w, \alpha\right)=\mathbbm{1}\left(\frac{1}{2} \notin \Theta_{\mathrm{CP}}\left(2 w \cdot \hat{\mu}_{w, \tau, j}, 2 w ; \alpha\right)\right),    \nonumber 
\end{equation}
where $w$ and $\alpha$ denote the window size and the significance level respectively, $\Theta_{\mathrm{CP}}$ denotes the CP interval and \begin{align}
\hat{\mu}_{w, \tau, j} = \frac{1}{2 w} \sum_{i = \tau-w+1}^{\tau}\left(\mathbbm{1}\left(\hat{y}_{i, j} = y_{i, j}\right)+\mathbbm{1}\left(\hat{y}_{i, j}^{\prime} = y_{i}^{\prime}\right)\right) .\nonumber
\end{align}
\par
The detector $d_{j}$ of the $j$-th two-sample test layer can directly determine whether a sample is ID or OOD for the $j$-th task, eliminating the need for a threshold. If the result is rejection and the $(j+1)$-th layer exists, the process proceeds to the next layer. If no further layers are present, it indicates that the sample has been rejected by all tests of $T$ layers. Consequently, we classify the new sample $x_{\tau}^{\prime}$ as OOD. Otherwise, the detection process terminates at the current layer through an early-exit mechanism. Since OOD for a specific task is not necessarily OOD for all tasks, whereas the ID of a task must also be ID for all tasks, we can classify $x_{\tau}^{\prime}$ as ID. Moreover, since each layer is task-specific, we can further infer that the task-id $t$ of the sample $x_{\tau}^{\prime}$ is $j$. As the Continual OOD Detection process described in Section \ref{sub_sec:cood}, after getting OOD and task-id predictions, ID samples' labels can be further predicted with their task-ids, and OOD samples can be used as training data for the new task once labeled.

Existing hierarchical testing frameworks primarily focus on hierarchical classification using predefined label hierarchies \cite{tieppo2022hierarchical}, however our work pioneers hierarchical OOD detection in open-world continual learning without relying on predefined label hierarchies. Specifically, we introduce a hierarchical architecture where each level focuses on distinct tasks. The hierarchical architecture achieves computational efficiency and enhanced performance through two fundamental mechanisms: (1) reduction in necessary detection operations, and (2) progressive simplification of decision boundaries as layers deepen. While H2ST and C2ST have the same number of classifiers, unlike C2ST must process all $T$ classifiers, H2ST avoids processing all classifiers for every sample. For a system with $T$ ID tasks under uniform sample distribution, C2ST necessitates evaluation of all source-target classifiers, requiring $T$ detection operations per sample. In contrast, H2ST, leveraging its hierarchical architecture and early-exit mechanism, is expected to perform only $\frac{T+1}{2}$ detection operations per sample. Furthermore, this architecture enables each classifier to focus on distinguishing only one ID task and fewer OOD tasks as the hierarchical architecture deepens, thereby streamlining the classification challenges faced by individual classifiers. Specifically, for the $j$-th layer, its OOD includes the $(j+1)$-th task to the $T$-th task, covering $T-j$ tasks, plus actual OOD for the entire system. As the hierarchical architecture deepens, $j$ increases, and the number of OOD tasks $T-j+1$ that the source-target classifier $g_{j}$ needs to distinguish decreases. In contrast, all classifiers in C2ST must handle $T$ OOD tasks. The hierarchical architecture fundamentally achieves better performance with lower overhead.

Besides the above benefits of the hierarchical architecture, our method offers the following advantages. First, the threshold-free method enables fully automated detection, enhancing adaptability to non-stationary data distributions in open-world scenarios. Moreover, H2ST leverages feature representations rather than raw data samples or model outputs, exploiting model capabilities while preventing over-dependence on model performance. Furthermore, H2ST achieves seamless integration with replay-based TIL methods by directly utilizing their inherent memory buffers.

\begin{algorithm}
\caption{Hierarchical Two-sample Tests (H2ST)}\label{alg:h2st}
\begin{algorithmic}[1]
\STATE \textbf{Input:}
\STATE \quad Window size \( w \)
\STATE \quad Significance level \( \alpha \)
\STATE \quad TIL model \( f_\theta \)
\STATE \quad Memory buffers \( \{B_1, B_2, \ldots, B_T\} \)
\STATE \quad Source-target classifiers \( \{g_1, g_2, \ldots, g_T\} \)
\FOR{each detection time \( \tau \)}
    \STATE Observe a target sample \( x_{\tau}^{\prime} \) and obtain its feature \( \psi(x_{\tau}^{\prime}) \) using \( f_\theta \)
    \FOR{each \( j \) in \( [1, T] \)}
        \STATE Draw source sample \( x_{\tau, j} \) from \( B_j \) to match the target quantity and obtain its feature \( \psi(x_{\tau, j}) \)
        \STATE Predict \( \hat{y}_{\tau, j} = g_j(\psi(x_{\tau, j})) \) and \( \hat{y}_{\tau, j}^{\prime} = g_j(\psi(x_{\tau}^{\prime})) \)
        \STATE Update \( g_j \) using source samples as 0 and target samples as 1, i.e., \( (x_{\tau, j}, 0) \) and \( (x_{\tau}^{\prime}, 1) \)
        \IF{ \( 0.5 \in \Theta_{\mathrm{CP}}(2w \cdot \hat{\mu}_{w, \tau, j}, 2w; \alpha) \) }
            \RETURN Detection result: \textit{ID}, task-id \( t \): \( j \)
        \ENDIF
    \ENDFOR
    \RETURN Detection result: \textit{OOD}
\ENDFOR
\end{algorithmic}
\end{algorithm}

\begin{table*}[t]
\small
\renewcommand{\arraystretch}{1.25}
%\scriptsize
  \centering
  \resizebox{\linewidth}{!}{
    \begin{tabular}{c|c|cc|cc|cc|cc|cc|cc|cc|cc}
    \hline
    &       & \multicolumn{14}{c}{Dataset}    &       &  \\
\cline{3-16}    TIL & OOD Method & \multicolumn{2}{c}{MNIST} & \multicolumn{2}{c}{SVHN} & \multicolumn{2}{c}{CIFAR-10} & \multicolumn{2}{c}{CIFAR-100} & \multicolumn{2}{c}{Mini-ImageNet} & \multicolumn{2}{c}{CoRe50} & \multicolumn{2}{c}{Stream-51} & \multicolumn{2}{c}{Average} \\
\cline{3-18}          &       & F1$\uparrow$    & TA$\uparrow$ & F1$\uparrow$    & TA$\uparrow$      & F1$\uparrow$    &TA$\uparrow$       & F1$\uparrow$    &TA$\uparrow$       & F1$\uparrow$    & TA$\uparrow$      & F1$\uparrow$   & TA$\uparrow$       & F1$\uparrow$   & TA$\uparrow$       & F1$\uparrow$   & TA$\uparrow$\\
    \hline
    \multirow{8}{*}{ER \cite{rolnick2019experience}}& MSP \cite{msp}   & 63.02  & 37.24  & 59.15  & 41.14  & 53.93  & 18.31  & 54.31  & 23.46  & 52.77  & 15.81  & 48.01  & 20.10  & 49.07  & 32.63  & 54.32  & 26.96  \\
    & Energy \cite{energy} & 60.79  & 37.69  & 61.77  & 49.01  & 54.38  & 19.90  & 54.35  & 22.91  & 52.81  & 16.20  & 50.41  & 34.50  & 53.72  & 44.67  & 55.46  & 32.13  \\
    & ODIN \cite{odin}  & 60.12  & 29.41  & 58.14  & 39.64  & 53.83  & 21.97  & 53.05  & 11.89  & 52.86  & 6.43  & 46.46  & 12.50  & 46.67  & 19.82  & 53.02  & 20.24  \\
    & Maxlogit \cite{maxlogit} &  62.53  & 40.18  & 65.03  & 54.83  & 53.97  & 20.53  & 54.38  & 19.56  & 53.00  & 13.80  & 50.83  & 33.47  & 55.24  & 46.85  & 56.43  & 32.74  \\
    & Gentropy \cite{GEN}   &  63.04  & 38.93  & 64.70  & 55.40  & 54.10  & 24.18  & 54.44  & 24.98  & 52.82  & 6.99  & 48.31  & 23.66  & 54.04  & 45.66  & 55.92  & 31.40  \\
    & FeatureNorm \cite{feature_norm} &  61.78  & 39.85  & 65.97  & 57.40  & 53.14  & 15.69  & 53.29  & 14.60  & 52.80  & 11.80  & 49.31  & 35.73  & 48.94  & 34.79  & 55.03  & 29.98  \\
    & MORE \cite{kim2022multi} & 70.30  & 68.10  & 71.91  & 69.00  & 62.97  & 58.80  & 56.11  & 68.92  & 55.25  & 69.54  & 68.15  & 70.39  & 67.12  & 64.30  & 64.54  & 67.01  \\
    & H2ST & \textbf{92.03} & \textbf{93.78} & \textbf{77.60} & \textbf{84.60} & \textbf{88.89} & \textbf{89.59} & \textbf{84.21} & \textbf{82.02} & \textbf{79.34} & \textbf{81.59} & \textbf{94.11} & \textbf{94.06} & \textbf{89.24} & \textbf{90.68} & \textbf{86.49} & \textbf{88.05} \\
    \hline
    \multirow{8}{*}{GEM \cite{gem}}& MSP \cite{msp}   &   63.07  & 38.88  & 60.66  & 43.41  & 53.83  & 20.03  & 54.22  & 20.80  & 52.97  & 11.91  & 48.19  & 23.68  & 48.83  & 31.50  & 54.54  & 27.17  \\
    & Energy \cite{energy} &64.64  & 41.73  & 63.01  & 51.24  & 54.26  & 22.41  & 54.44  & 27.62  & 53.00  & 11.65  & 48.75  & 27.16  & 49.74  & 31.13  & 55.41  & 30.42  \\
    & ODIN \cite{odin}  &  60.66  & 27.99  & 57.63  & 35.94  & 54.04  & 23.21  & 53.06  & 11.87  & 52.86  & 6.48  & 46.30  & 13.90  & 47.13  & 13.43  & 53.10  & 18.97  \\
    & Maxlogit \cite{maxlogit} & 63.61  & 40.76  & 64.32  & 54.82  & 54.25  & 24.24  & 54.35  & 22.47  & 53.02  & 13.71  & 50.99  & 39.36  & 52.17  & 39.72  & 56.10  & 33.58  \\
    & Gentropy \cite{GEN}   & 64.46  & 40.22  & 63.92  & 54.10  & 53.79  & 24.97  & 54.45  & 25.62  & 52.86  & 6.30  & 51.10  & 40.58  & 53.37  & 43.56  & 56.28  & 33.62  \\
    & FeatureNorm \cite{feature_norm} & 61.14  & 38.83  & 64.54  & 53.47  & 53.48  & 23.62  & 53.28  & 14.64  & 52.90  & 11.56  & 48.87  & 35.58  & 48.86  & 32.47  & 54.73  & 30.02  \\
    & MORE \cite{kim2022multi} & 71.95  & 69.68  & 72.62  & 71.03  & 66.54  & 62.72  & 56.63  & 69.19  & 53.28  & 68.60  & 70.69  & 71.37  & 69.52  & 68.33  & 65.89  & 68.70  \\
    & H2ST &  \textbf{91.55} & \textbf{93.43} & \textbf{77.87} & \textbf{85.20} & \textbf{93.54} & \textbf{92.82} & \textbf{89.98} & \textbf{84.37} & \textbf{83.88} & \textbf{82.69} & \textbf{95.38} & \textbf{93.94} & \textbf{91.63} & \textbf{91.23} & \textbf{89.12} & \textbf{89.10} \\
    \hline
    \end{tabular}
    }
  \caption{Comparison of the OOD detection performance across different methods.}
  \label{tab:main_ood}
\end{table*}
\section{Experiments}
\label{sec:experiments}
\subsection{Experimental Setup}
\label{sub_sec:exp_setup}
\textbf{Dataset.} We evaluate on the following datasets: MNIST \cite{mnist}, SVHN \cite{netzer2011svhn}, CIFAR-10 \cite{krizhevsky2009cifar}, CIFAR-100 \cite{krizhevsky2009cifar}, Mini-ImageNet \cite{vinyals2016minii}, CoRe50 \cite{lomonaco2017core50}, and Stream-51 \cite{Roady_2020_Stream51}. For MNIST, SVHN, and CIFAR-10, we construct five tasks with two classes each; for CIFAR-100 and Mini-ImageNet, five tasks with twenty classes each. The data stream is constructed in a 6k-2k-2k configuration: when a task serves as OOD, it contains 6,000 samples, followed by two OOD detection rounds, where 2,000 samples are used as ID to facilitate a smooth transition. CoRe50 and Stream-51 use ten tasks with five classes each in a 9k-2k-2k configuration.
\par \noindent
\textbf{TIL Methods.} 
For a fair comparison, we adopt GEM \cite{gem} and Experience Replay (ER) \cite{rolnick2019experience} with class-balanced memory management among all detection methods. Notably, our H2ST is compatible with any replay-based TIL method.
\par \noindent
\textbf{Model Architectures.} 
For MNIST, we use a two-layer fully connected network (100 ReLU units per layer). For the other tasks, we employ a smaller version of ResNet18 \cite{resnet}. A fully connected neural network with a single hidden layer of 128 ReLU units serves as the source-target classifier.
\par \noindent
\textbf{OOD Detection Baselines.} We select six OOD detection methods and one OOD-based CL method as baselines, including maximum softmax probability (MSP) \cite{msp}, Energy score \cite{energy}, ODIN \cite{odin}, MaxLogit \cite{maxlogit}, Gentropy \cite{GEN}, FeatureNorm \cite{feature_norm}, and MORE \cite{kim2022multi}. For methods that require a threshold to distinguish ID from OOD samples, we determine optimal thresholds via an iterative search from minimum to maximum values in steps of decile/100. Notably, this threshold assumes that labels are known, which is impractical in real-world applications. We traverse each ID task and OOD separately to obtain respective thresholds. If a sample meets multiple thresholds, task-id assignment becomes infeasible.
\par \noindent
\textbf{Evaluation Metrics.} For comparison, we select four metrics to evaluate both classification and OOD detection. For classification, ACC is the average accuracy across all tasks, while forgetting (FT) measures how learning new tasks affects previous tasks. For OOD detection, F1 indicates the average F1 score of each detection, and TA denotes the average accuracy of task-id prediction. TA is a more stringent metric than F1, as F1 requires only binary classification between ID and OOD, whereas TA additionally requires identifying the specific task to which an ID sample belongs. All metrics are reported as percentages.
\begin{table*}[htbp]
\small
\renewcommand{\arraystretch}{1.25}
  \centering
  \resizebox{\linewidth}{!}{
    \begin{tabular}{c|c|cc|cc|cc|cc|cc|cc|cc|cc}
    \hline
    &       & \multicolumn{14}{c}{Dataset}    &       &  \\
\cline{3-16}    TIL & OOD Method & \multicolumn{2}{c}{MNIST} & \multicolumn{2}{c}{SVHN} & \multicolumn{2}{c}{CIFAR-10} & \multicolumn{2}{c}{CIFAR-100} & \multicolumn{2}{c}{Mini-ImageNet} & \multicolumn{2}{c}{CoRe50} & \multicolumn{2}{c}{Stream-51} & \multicolumn{2}{c}{Average} \\
\cline{3-18}          &       & ACC$\uparrow$    & FT$\uparrow$ & ACC$\uparrow$    & FT$\uparrow$      & ACC$\uparrow$    &FT$\uparrow$       & ACC$\uparrow$    &FT$\uparrow$       & ACC$\uparrow$    & FT$\uparrow$      & ACC$\uparrow$   & FT$\uparrow$       & ACC$\uparrow$   & FT$\uparrow$       & ACC$\uparrow$   & FT$\uparrow$\\
    \hline
    \multirow{8}{*}{ER \cite{rolnick2019experience}}& MSP \cite{msp}  & 96.24  & -1.23  & 93.46  & -3.74  & 67.49  & \textbf{5.20 } & 24.03  & -3.81  & 19.49  & -6.28  & 68.18  & \textbf{10.75} & 65.42  & 4.81  & 62.05  & 0.81  \\
    & Energy \cite{energy}& 77.65  & -0.26  & \textbf{95.42} & \textbf{-1.42} & 62.87  & -2.91  & 22.60  & -1.06  & 17.38  & -3.22  & 65.83  & -0.79  & 69.18  & -1.80  & 58.70  & -1.64  \\
    & ODIN \cite{odin} &  91.63  & \textbf{0.45} & 93.57  & -3.26  & 78.02  & 0.33  & 19.23  & -2.11  & 12.82  & \textbf{0.74} & 49.22  & 9.35  & 59.35  & \textbf{7.65} & 57.69  & \textbf{1.88} \\
    & Maxlogit \cite{maxlogit} & 95.75  & -1.39  & 94.08  & -3.09  & 67.00  & -1.93  & 22.10  & -1.20  & 13.80  & -1.68  & 68.61  & 7.07  & 63.02  & -12.80  & 60.62  & -2.14  \\
    & Gentropy \cite{GEN}   & 96.14  & -1.73  & 95.19  & -1.74  & 67.37  & -5.73  & 25.17  & -2.18  & 12.34  & -1.71  & 49.96  & 4.93  & 69.52  & -7.34  & 59.38  & -2.21  \\
    & FeatureNorm \cite{feature_norm} & 97.28  & -0.44  & 94.92  & -2.09  & 61.79  & -2.45  & 21.91  & \textbf{1.80} & 17.26  & -6.85  & 57.86  & -5.57  & 53.51  & -0.20  & 57.79  & -2.26  \\
    & MORE \cite{kim2022multi} & 97.89  & -0.49  & 93.77  & -2.13  & 80.16  & -10.61  & 43.92  & -14.39  & 30.31  & -11.28  & 66.78  & -1.55  & 68.05  & -4.46  & 68.70  & -6.41  \\
    & H2ST &  \textbf{98.49} & -0.80  & 93.45  & -4.24  & \textbf{84.34} & -8.98  & \textbf{45.09} & -14.23  & \textbf{31.91} & -10.56  & \textbf{78.26} & -1.42  & \textbf{74.33} & -5.50  & \textbf{72.27} & -6.53  \\

    \hline
    \multirow{8}{*}{GEM \cite{gem}}& MSP \cite{msp}&   95.14  & \textbf{2.65} & 95.24  & -1.19  & 71.18  & -1.34  & 27.27  & \textbf{2.75} & 15.60  & -3.68  & 64.62  & \textbf{8.38} & 57.86  & 1.50  & 60.99  & \textbf{1.30} \\
    & Energy \cite{energy}& 96.71  & -0.02  & 94.85  & -2.71  & 77.52  & \textbf{1.34} & 35.38  & -6.49  & 16.41  & -5.50  & 60.88  & 6.80  & 55.15  & \textbf{3.18} & 62.41  & -0.49  \\
    & ODIN \cite{odin}& 77.95  & -0.95  & 72.92  & \textbf{3.24} & 67.62  & -4.84  & 21.51  & -0.65  & 13.07  & \textbf{-0.10} & 44.44  & 10.14  & 37.31  & -4.93  & 47.83  & 0.27  \\
    & Maxlogit \cite{maxlogit}& 97.96  & -0.79  & 94.79  & -2.37  & 65.08  & -6.33  & 22.70  & -1.83  & 16.14  & -3.56  & 60.73  & -5.53  & 58.28  & -0.19  & 59.38  & -2.94  \\
    & Gentropy \cite{GEN}& 97.95  & -0.46  & \textbf{95.45} & -1.61  & 70.95  & -3.48  & 25.19  & -2.44  & 11.98  & -0.21  & 67.95  & 0.09  & 64.74  & -3.65  & 62.03  & -1.68  \\
    & FeatureNorm \cite{feature_norm}& 96.58  & -0.15  & 92.68  & -4.64  & 62.11  & -2.35  & 20.38  & -0.15  & 16.40  & -2.30  & 62.51  & 1.14  & 54.34  & 1.52  & 57.86  & -0.99  \\
    & MORE \cite{kim2022multi}& 98.32  & -0.55  & 93.08  & -3.15  & 83.61  & -6.59  & 42.95  & -15.30  & \textbf{31.56} & -9.98  & 70.97  & 2.10  & 68.31  & -1.10  & 69.83  & -4.94  \\
    & H2ST & \textbf{98.43} & -1.01  & 92.64  & -5.20  & \textbf{84.71} & -7.90  & \textbf{45.91} & -13.90  & 30.80  & -12.24  & \textbf{78.31} & -1.64  & \textbf{69.36} & -13.13  & \textbf{71.45} & -7.86  \\
    \hline
    \end{tabular}
    }
  \caption{Comparison of the TIL performance across different methods.}
  \label{tab:main_til}
\end{table*}
\subsection{Main Results}
\label{sub_sec:main_res}
\textbf{OOD Detection Results.} Table \ref{tab:main_ood} provides a comprehensive evaluation of OOD detection performance in terms of average F1 score and average task-id prediction accuracy for each method. It's evident that our H2ST consistently and significantly outperforms all other OOD detection methods across different cases. Notably, on the CIFAR-10 dataset, H2ST achieves an increase in average F1 of 25.92\% and average TA of 30.79\% over the second-best method using ER. The CIFAR-100 and Mini-ImageNet datasets present a higher level of complexity due to the increased number of classes per task, and H2ST achieves F1 scores of 84.21\% and 79.34\% and TA of 82.02\% and 81.59\%. On CoRe50 and Stream-51 with larger sample sizes, H2ST shows robust performance with F1 scores of 94.11\% and 89.24\% and TA of 94.06\% and 90.68\%, surpassing all baselines. Furthermore, H2ST demonstrates strong generalization capabilities across continual learning frameworks. When integrated with the GEM framework, H2ST maintains its superior performance, achieving average F1 of 89.12\% and TA of 89.10\%, outperforming the second-best method by 23.23\% and 20.40\%, respectively. These results underscore H2ST's effectiveness and pronounced advantage in continual OOD detection.
\par \noindent
\textbf{TIL Results.} Table \ref{tab:main_til} presents the TIL performance. It is essential to clarify that continual learning itself is not the primary focus of our proposed H2ST, as H2ST is fundamentally an OOD detection method designed for continual learning scenarios, tailored explicitly to TIL frameworks. In open-world TIL, models utilize detected OOD samples as training data for new tasks, thereby establishing a direct dependency between prediction performance and the efficacy of OOD detection. As indicated in Table \ref{tab:main_ood}, H2ST's superior OOD detection capability enables it to identify more OOD samples. This enhanced detection enables the model to access more diverse training data for each task, leading to average ACC increases by 2.59\% over the second-best method. Meanwhile, it concurrently exacerbates forgetting. Less detected OOD samples lead to less new task acquisition, and can mitigate forgetting, but conflict with the objective of continual learning. This tension highlights a critical trade-off inherent to continual learning: balancing new knowledge integration against previous knowledge preservation, but it's not the focus of this paper.
\subsection{Ablation Studies}
\label{sub_sec:ablation}
\textbf{Better Performance with Lower Overhead than C2ST.}
In Section \ref{sec:method}, we theoretically analyze the advantages of the hierarchical architecture compared with C2ST in terms of performance and overhead. Now, we provide experimental results to further validate these benefits. As shown in Fig. \ref{fig:parallel}, H2ST and C2ST exhibit comparable performance in TIL. Detailed results are provided in the appendix. For OOD detection, however, the hierarchical architecture demonstrates considerable improvement, with an average increase of 11.37\% in F1 and 4.20\% in TA across all datasets and TIL methods. We visualize results for each detection as the number of tasks ($i.e.$, the hierarchical depth of H2ST) increases on CoRe50 in Fig. \ref{fig:depth}. More visualization results are provided in the appendix. Our findings demonstrate that H2ST delivers stable and superior detection performance than C2ST. We also compare the time different methods need to process each input. Specifically, when $T$=9, H2ST processes each input in 16.7 ms, including all necessary source-target classifier prediction and update, compared to C2ST's average of 22.0 ms. Overall, H2ST's hierarchical architecture achieves superior detection performance with less overhead.  It should be noted that we compare the performance of H2ST with baselines in Table \ref{tab:main_ood}, but omit computational overhead comparisons. Unlike traditional score-based OOD methods, H2ST and C2ST require additional source-target classifiers, which incur extra overhead, but as shown in the table, this overhead is justified by a substantial improvement in detection performance. For the single-C2ST that cannot predict task-id, experimental results are provided in the appendix.
\begin{figure}[t]
  \centering
   \includegraphics[width=\linewidth]{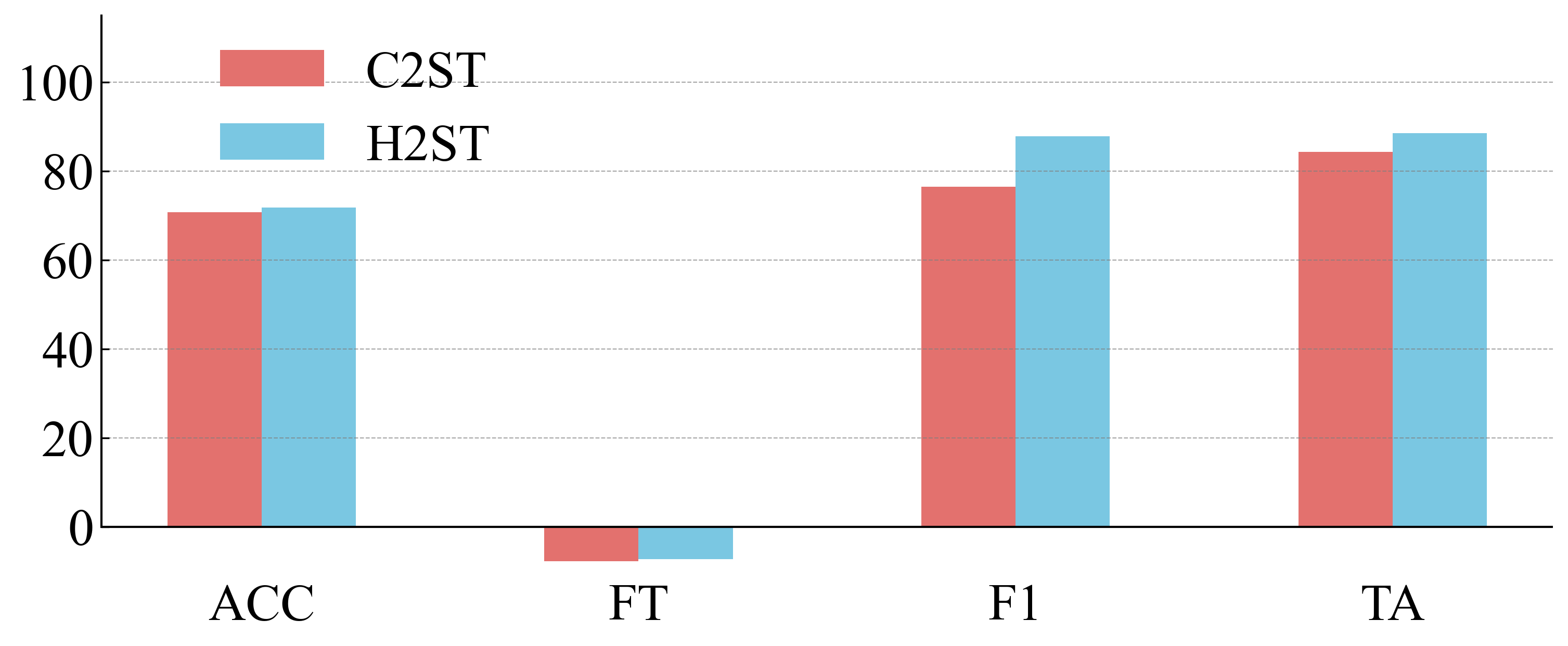}
   \caption{Performance comparison between H2ST and C2ST. Both exhibit comparable TIL performance, but the hierarchical architecture demonstrates superior OOD detection performance.}
   \label{fig:parallel}
\end{figure}

\begin{figure}[t]
  \centering
   \includegraphics[width=\linewidth]{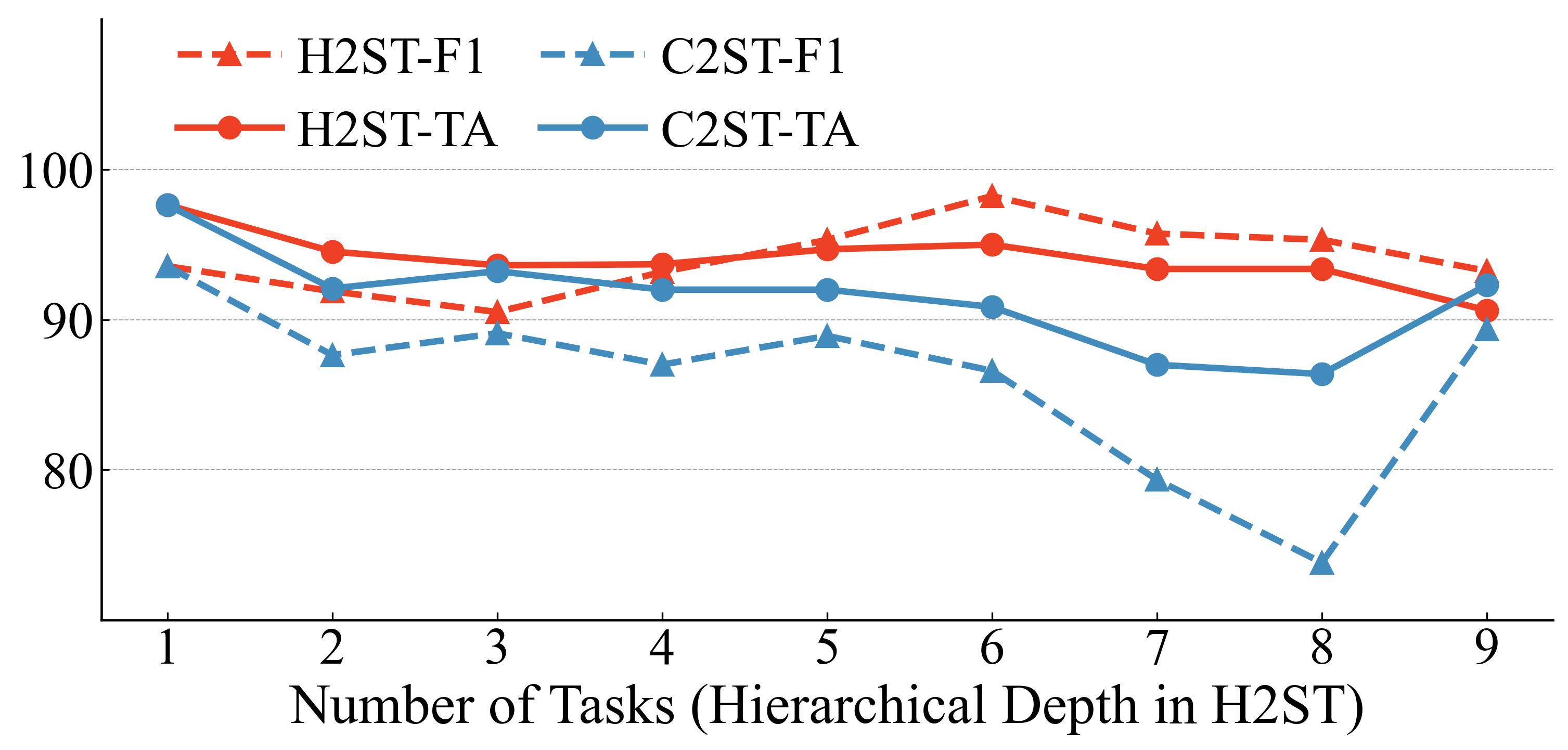}
   \caption{Performance sensitivity to depth. As tasks scale, H2ST delivers stable and superior detection performance.}
   \label{fig:depth}
\end{figure}
\par
\begin{figure}[t]
  \centering
   \includegraphics[width=\linewidth]{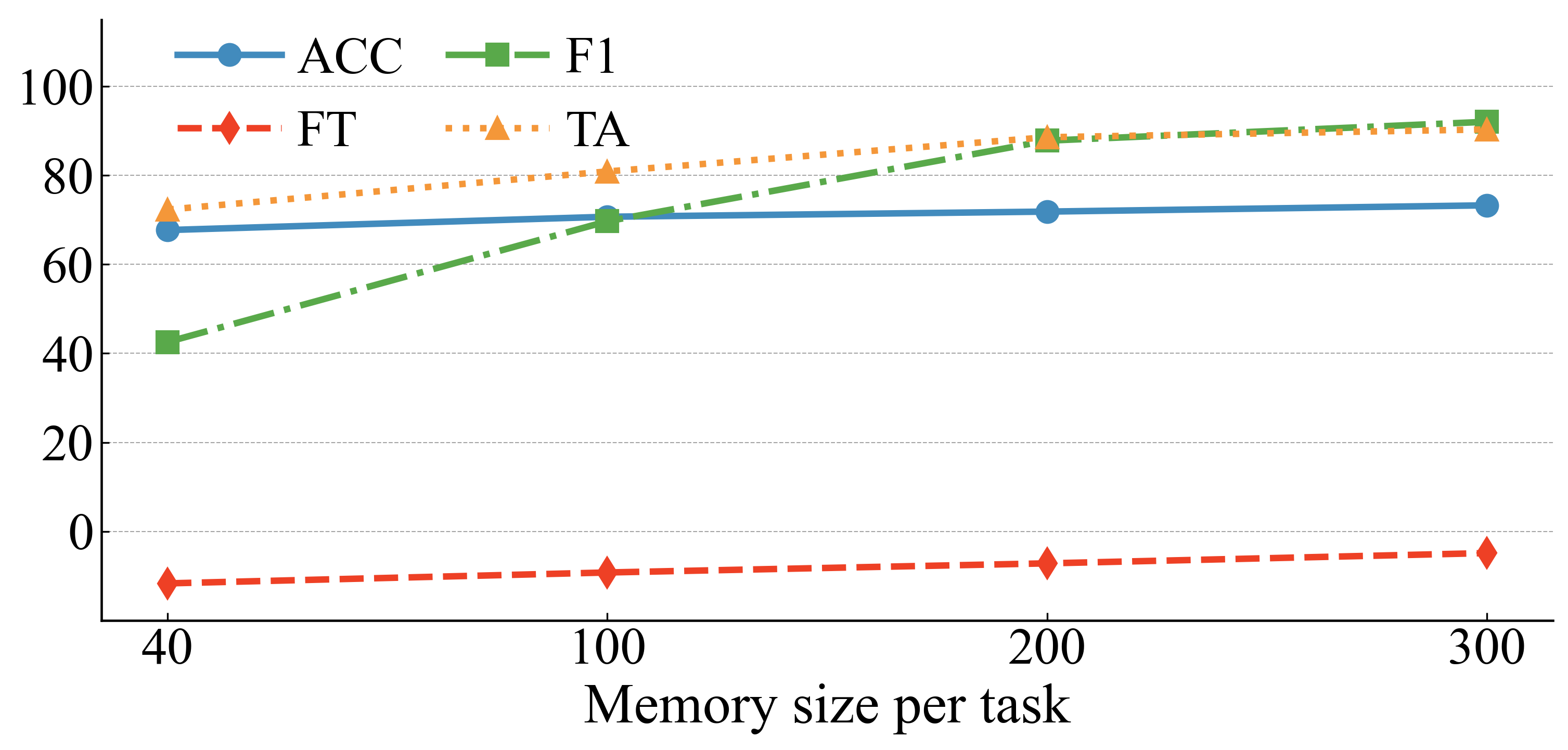}
   \caption{Impact of memory size on performance metrics. All metrics improve with memory size, though OOD detection shows diminishing returns at larger sizes.}
   \label{fig:memory}
\end{figure}

 \noindent
\textbf{Analysis of Memory Size.} The memory size is crucial in replay-based TIL methods. A larger memory enhances the model's capacity to retain prior knowledge and offers a richer representation of learned task distributions. However, it demands more storage and computational overhead. In H2ST, random samples are drawn from memory to serve as source samples, making performance potentially sensitive to memory size. Fig. \ref{fig:memory} illustrates the trend of metrics on average across all datasets and TIL methods relative to memory size. Detailed results are provided in the appendix. All metrics improve with increasing memory size. The increases in ACC and FT directly result from more stored samples. Regarding OOD detection, increases arise for several reasons. First, better model performance and reduced forgetting enable more effective feature representation. Second, sufficient memory size reduces the risk of biased representation, ensuring randomly drawn samples represent the task distribution. However, once beyond a certain extent, increasing memory size further results in diminishing returns. Overall, determining an optimal memory size requires careful consideration of the trade-off between achieving peak performance and cost-effectiveness. Moreover, Table \ref{tab:memory} presents comparative results of different methods under different memory sizes on the CIFAR-10 dataset, quantitatively confirming H2ST's performance superiority across all memory sizes.

\begin{table}[t]
  \centering
  \small
   \resizebox{\linewidth}{!}{
    \begin{tabular}{c|ccc|ccc|ccc}
    \hline
    \multirow{2}{*}{Size} & \multicolumn{3}{c}{H2ST} & \multicolumn{3}{c}{C2ST} & \multicolumn{3}{c}{Gentropy} \\
\cline{2-10}          & ACC$\uparrow$   & F1$\uparrow$    & TA$\uparrow$    & ACC$\uparrow$   & F1$\uparrow$    & TA$\uparrow$    & ACC$\uparrow$   & F1$\uparrow$    & TA$\uparrow$ \\
    \hline
    40    & \textbf{74.78} & \textbf{57.41} & \textbf{75.13} & 74.76  & 53.28  & 74.58  & 67.61  & 53.70  & 26.39  \\
    100   & 78.72  & \textbf{78.57} & \textbf{84.26} & \textbf{80.09} & 73.56  & 81.26  & 59.72  & 53.79  & 25.21  \\
    200   & \textbf{84.71} & \textbf{93.54} & \textbf{92.82} & 84.41  & 87.88  & 89.35  & 70.95  & 53.79  & 24.97  \\
    300   & \textbf{85.37} & \textbf{94.41} & \textbf{93.01} & 84.92  & 90.34  & 91.58  & 78.63  & 54.12  & 28.47  \\
    \bottomrule
    \end{tabular}
    }
    \caption{Performance comparison across memory sizes. H2ST's overall performance surpasses baselines across all memory sizes.}
  \label{tab:memory}
\end{table}

\begin{table}[t]
  \centering
  \small
  \resizebox{\linewidth}{!}{
    \begin{tabular}{c|cccc|cccc}
    \hline
\multirow{2}{*}{Data Stream} & \multicolumn{4}{c}{CIFAR-10}  & \multicolumn{4}{c}{CIFAR-100} \\
\cline{2-9}          & ACC$\uparrow$   & FT$\uparrow$    & F1$\uparrow$    & TA$\uparrow$    & ACC$\uparrow$   & FT$\uparrow$    & F1$\uparrow$    & TA$\uparrow$ \\
    \hline
    2k-4k-4k & 81.83  & \textbf{-5.07} & 83.29  & 75.15  & 34.23  & \textbf{-8.42} & 52.94  & 42.32  \\
    4k-3k-3k & 84.02  & -6.67  & 90.35  & 85.74  & 39.97  & -10.89  & 68.18  & 65.40  \\
    6k-2k-2k & \textbf{84.53} & -8.44  & 91.22  & 91.21  & 45.50  & -14.06  & 87.09  & 83.19  \\
    8k-1k-1k & 84.47  & -9.34  & \textbf{93.34} & \textbf{95.69} & \textbf{46.93} & -15.45  & \textbf{91.32} & \textbf{89.30} \\
    \hline
    \end{tabular}
    }
    \caption{Performance comparison across data streams. As test OOD samples increase, more samples are detected and used as training samples for new tasks, resulting in improved performance.} 
  \label{tab:stream}
\end{table}
\par \noindent
\textbf{Analysis of Data streams.} We conducted comprehensive experiments across various data stream configurations (ranging from 2k-4k-4k to 8k-1k-1k) using both ER and GEM, with average performance metrics provided in Table \ref{tab:stream}. In the 2k-4k-4k configuration, a task serving as OOD contains 2,000 samples, while the subsequent two OOD detection rounds each use 4,000 ID samples. As the number of OOD test samples increases from 2,000 to 8,000, the average FT decreases from -6.74\% to -12.40\%, which can be attributed to the increased number of training data of new tasks. Meanwhile, on average, ACC improves by 7.67\%, F1 by 24.21\%, and TA by 33.77\%, indicating better OOD detection performance with more OOD test samples. First, as the source-target classifiers are updated online with each detection, more test samples expose them to more diverse data. Second, if the OOD detection performance is sufficiently good, more OOD samples during testing could mean more detected OOD samples, resulting in more training data for TIL and therefore better model performance. This enhancement in model performance in turn positively impacts OOD detection. In continual OOD detection, continual learning and OOD detection are interrelated processes that influence one another. To enable the TIL model to autonomously detect and learn new tasks in open-world, achieving strong performance in both TIL and OOD detection is essential.

\section{Conclusion}
\label{sec:conclu}
\par
In this paper, we introduce open-world TIL scenario, wherein the model needs to detect OOD samples and predict task-ids and labels. To address this challenging setting, we propose Hierarchical Two-Sample Tests (H2ST). H2ST employs a hierarchical architecture that allows for efficient and fine-grained detection without threshold selection, exploits model capabilities without over-dependence, and achieves seamless integration with replay-based TIL methods.
Extensive experiments demonstrate the effectiveness and superiority of H2ST in continual OOD detection.

{
    \small
    \bibliographystyle{ieeenat_fullname}
    \bibliography{main}

\begin{thebibliography}{57}
\providecommand{\natexlab}[1]{#1}
\providecommand{\url}[1]{\texttt{#1}}
\expandafter\ifx\csname urlstyle\endcsname\relax
  \providecommand{\doi}[1]{doi: #1}\else
  \providecommand{\doi}{doi: \begingroup \urlstyle{rm}\Url}\fi

\bibitem[Aguilar et~al.(2023)Aguilar, Raducanu, Radeva, and Van~de Weijer]{aguilar2023cedl}
Eduardo Aguilar, Bogdan Raducanu, Petia Radeva, and Joost Van~de Weijer.
\newblock Continual evidential deep learning for out-of-distribution detection.
\newblock In \emph{Proceedings of the IEEE/CVF International Conference on Computer Vision}, pages 3444--3454, 2023.

\bibitem[Aljundi et~al.(2017)Aljundi, Chakravarty, and Tuytelaars]{aljundi2017expert}
Rahaf Aljundi, Punarjay Chakravarty, and Tinne Tuytelaars.
\newblock Expert gate: Lifelong learning with a network of experts.
\newblock In \emph{Proceedings of the IEEE conference on computer vision and pattern recognition}, pages 3366--3375, 2017.

\bibitem[Aljundi et~al.(2018)Aljundi, Babiloni, Elhoseiny, Rohrbach, and Tuytelaars]{aljundi2018memoryAS}
Rahaf Aljundi, Francesca Babiloni, Mohamed Elhoseiny, Marcus Rohrbach, and Tinne Tuytelaars.
\newblock Memory aware synapses: Learning what (not) to forget.
\newblock In \emph{Proceedings of the European conference on computer vision (ECCV)}, pages 139--154, 2018.

\bibitem[Aljundi et~al.(2022)Aljundi, Reino, Chumerin, and Turner]{aljundi2022continualnovelty}
Rahaf Aljundi, Daniel~Olmeda Reino, Nikolay Chumerin, and Richard~E Turner.
\newblock Continual novelty detection.
\newblock In \emph{Conference on Lifelong Learning Agents}, pages 1004--1025. PMLR, 2022.

\bibitem[Bang et~al.(2021)Bang, Kim, Yoo, Ha, and Choi]{bang2021rainbow}
Jihwan Bang, Heesu Kim, YoungJoon Yoo, Jung-Woo Ha, and Jonghyun Choi.
\newblock Rainbow memory: Continual learning with a memory of diverse samples.
\newblock In \emph{Proceedings of the IEEE/CVF conference on computer vision and pattern recognition}, pages 8218--8227, 2021.

\bibitem[Brown et~al.(2001)Brown, Cai, and DasGupta]{brown2001interval}
Lawrence~D Brown, T~Tony Cai, and Anirban DasGupta.
\newblock Interval estimation for a binomial proportion.
\newblock \emph{Statistical science}, 16\penalty0 (2):\penalty0 101--133, 2001.

\bibitem[Chen and Liu(2018)]{chen2018lifelong}
Zhiyuan Chen and Bing Liu.
\newblock \emph{Lifelong machine learning}.
\newblock Morgan \& Claypool Publishers, 2018.

\bibitem[De~Lange et~al.(2021)De~Lange, Aljundi, Masana, Parisot, Jia, Leonardis, Slabaugh, and Tuytelaars]{de2021cl}
Matthias De~Lange, Rahaf Aljundi, Marc Masana, Sarah Parisot, Xu Jia, Ale{\v{s}} Leonardis, Gregory Slabaugh, and Tinne Tuytelaars.
\newblock A continual learning survey: Defying forgetting in classification tasks.
\newblock \emph{IEEE transactions on pattern analysis and machine intelligence}, 44\penalty0 (7):\penalty0 3366--3385, 2021.

\bibitem[Deng(2012)]{mnist}
Li Deng.
\newblock The mnist database of handwritten digit images for machine learning research.
\newblock \emph{IEEE Signal Processing Magazine}, 29\penalty0 (6):\penalty0 141--142, 2012.

\bibitem[Dong et~al.(2022)Dong, Guo, Li, Ting, Liu, and Kung]{dong2022nmd}
Xin Dong, Junfeng Guo, Ang Li, Wei-Te Ting, Cong Liu, and HT Kung.
\newblock Neural mean discrepancy for efficient out-of-distribution detection.
\newblock In \emph{Proceedings of the IEEE/CVF Conference on Computer Vision and Pattern Recognition}, pages 19217--19227, 2022.

\bibitem[Drummond and Shearer(2006)]{open_world}
Nick Drummond and Rob Shearer.
\newblock The open world assumption.
\newblock In \emph{eSI Workshop: The Closed World of Databases meets the Open World of the Semantic Web}, page~1, 2006.

\bibitem[Evron et~al.(2022)Evron, Moroshko, Ward, Srebro, and Soudry]{evron2022cf}
Itay Evron, Edward Moroshko, Rachel Ward, Nathan Srebro, and Daniel Soudry.
\newblock How catastrophic can catastrophic forgetting be in linear regression?
\newblock In \emph{Conference on Learning Theory}, pages 4028--4079. PMLR, 2022.

\bibitem[Friedman(2003)]{friedman2003multivariate}
Jerome~H Friedman.
\newblock On multivariate goodness of fit and two sample testing. econf.
\newblock \emph{C030908: THPD002}, 2003.

\bibitem[Glorot and Bengio(2010)]{glorot2010understanding}
Xavier Glorot and Yoshua Bengio.
\newblock Understanding the difficulty of training deep feedforward neural networks.
\newblock In \emph{Proceedings of the thirteenth international conference on artificial intelligence and statistics}, pages 249--256. JMLR Workshop and Conference Proceedings, 2010.

\bibitem[Guo et~al.(2020)Guo, Liu, Yang, and Rosing]{guo2020improved}
Yunhui Guo, Mingrui Liu, Tianbao Yang, and Tajana Rosing.
\newblock Improved schemes for episodic memory-based lifelong learning.
\newblock \emph{Advances in Neural Information Processing Systems}, 33:\penalty0 1023--1035, 2020.

\bibitem[Hartley and Fitch(1951)]{hartley1951chart}
HO Hartley and ER Fitch.
\newblock A chart for the incomplete beta-function and the cumulative binomial distribution.
\newblock \emph{Biometrika}, 38\penalty0 (3/4):\penalty0 423--426, 1951.

\bibitem[He and Zhu(2022)]{he2022out}
Jiangpeng He and Fengqing Zhu.
\newblock Out-of-distribution detection in unsupervised continual learning.
\newblock In \emph{Proceedings of the IEEE/CVF Conference on Computer Vision and Pattern Recognition}, pages 3850--3855, 2022.

\bibitem[He et~al.(2016)He, Zhang, Ren, and Sun]{resnet}
Kaiming He, Xiangyu Zhang, Shaoqing Ren, and Jian Sun.
\newblock Deep residual learning for image recognition.
\newblock In \emph{Proceedings of the IEEE conference on computer vision and pattern recognition}, pages 770--778, 2016.

\bibitem[Hendrycks and Gimpel(2018)]{msp}
Dan Hendrycks and Kevin Gimpel.
\newblock A baseline for detecting misclassified and out-of-distribution examples in neural networks, 2018.

\bibitem[Hendrycks et~al.(2022)Hendrycks, Basart, Mazeika, Zou, Kwon, Mostajabi, Steinhardt, and Song]{maxlogit}
Dan Hendrycks, Steven Basart, Mantas Mazeika, Andy Zou, Joe Kwon, Mohammadreza Mostajabi, Jacob Steinhardt, and Dawn Song.
\newblock Scaling out-of-distribution detection for real-world settings.
\newblock \emph{ICML}, 2022.

\bibitem[Hossain et~al.(2022)Hossain, Saha, Chowdhury, Rahman, Rahman, and Mohammed]{hossain2022rethinkingtil}
Md~Sazzad Hossain, Pritom Saha, Townim~Faisal Chowdhury, Shafin Rahman, Fuad Rahman, and Nabeel Mohammed.
\newblock Rethinking task-incremental learning baselines.
\newblock In \emph{2022 26th International Conference on Pattern Recognition (ICPR)}, pages 2771--2777. IEEE, 2022.

\bibitem[Huang et~al.(2021)Huang, Geng, and Li]{huang2021gradnorm}
Rui Huang, Andrew Geng, and Yixuan Li.
\newblock On the importance of gradients for detecting distributional shifts in the wild.
\newblock \emph{Advances in Neural Information Processing Systems}, 34:\penalty0 677--689, 2021.

\bibitem[Hyder et~al.(2022)Hyder, Shao, Hou, Markopoulos, Prater-Bennette, and Asif]{hyder2022incremental}
Rakib Hyder, Ken Shao, Boyu Hou, Panos Markopoulos, Ashley Prater-Bennette, and M~Salman Asif.
\newblock Incremental task learning with incremental rank updates.
\newblock In \emph{European Conference on Computer Vision}, pages 566--582. Springer, 2022.

\bibitem[Jang et~al.(2022)Jang, Park, Lee, and Bastani]{jang2022sequential}
Sooyong Jang, Sangdon Park, Insup Lee, and Osbert Bastani.
\newblock Sequential covariate shift detection using classifier two-sample tests.
\newblock In \emph{International Conference on Machine Learning}, pages 9845--9880. PMLR, 2022.

\bibitem[Jiang and Celiktutan(2023)]{jiang2023neural}
Jian Jiang and Oya Celiktutan.
\newblock Neural weight search for scalable task incremental learning.
\newblock In \emph{Proceedings of the IEEE/CVF Winter Conference on Applications of Computer Vision}, pages 1390--1399, 2023.

\bibitem[Kang et~al.(2022)Kang, Mina, Madjid, Yoon, Hasegawa-Johnson, Hwang, and Yoo]{kang2022forgetWSN}
Haeyong Kang, Rusty John~Lloyd Mina, Sultan Rizky~Hikmawan Madjid, Jaehong Yoon, Mark Hasegawa-Johnson, Sung~Ju Hwang, and Chang~D Yoo.
\newblock Forget-free continual learning with winning subnetworks.
\newblock In \emph{International Conference on Machine Learning}, pages 10734--10750. PMLR, 2022.

\bibitem[Kim et~al.(2022{\natexlab{a}})Kim, Liu, and Ke]{kim2022multi}
Gyuhak Kim, Bing Liu, and Zixuan Ke.
\newblock A multi-head model for continual learning via out-of-distribution replay.
\newblock In \emph{Conference on Lifelong Learning Agents}, pages 548--563. PMLR, 2022{\natexlab{a}}.

\bibitem[Kim et~al.(2022{\natexlab{b}})Kim, Xiao, Konishi, Ke, and Liu]{kim2022theoretical}
Gyuhak Kim, Changnan Xiao, Tatsuya Konishi, Zixuan Ke, and Bing Liu.
\newblock A theoretical study on solving continual learning.
\newblock \emph{Advances in neural information processing systems}, 35:\penalty0 5065--5079, 2022{\natexlab{b}}.

\bibitem[Kim et~al.(2024)Kim, Xiao, Konishi, Ke, and Liu]{kim2024OCL}
Gyuhak Kim, Changnan Xiao, Tatsuya Konishi, Zixuan Ke, and Bing Liu.
\newblock Open-world continual learning: Unifying novelty detection and continual learning.
\newblock \emph{Artificial Intelligence}, page 104237, 2024.

\bibitem[Kirkpatrick et~al.(2017)Kirkpatrick, Pascanu, Rabinowitz, Veness, Desjardins, Rusu, Milan, Quan, Ramalho, Grabska-Barwinska, et~al.]{kirkpatrick2017overcoming}
James Kirkpatrick, Razvan Pascanu, Neil Rabinowitz, Joel Veness, Guillaume Desjardins, Andrei~A Rusu, Kieran Milan, John Quan, Tiago Ramalho, Agnieszka Grabska-Barwinska, et~al.
\newblock Overcoming catastrophic forgetting in neural networks.
\newblock \emph{Proceedings of the national academy of sciences}, 114\penalty0 (13):\penalty0 3521--3526, 2017.

\bibitem[Krizhevsky et~al.(2009)Krizhevsky, Hinton, et~al.]{krizhevsky2009cifar}
Alex Krizhevsky, Geoffrey Hinton, et~al.
\newblock Learning multiple layers of features from tiny images.
\newblock 2009.

\bibitem[Liang et~al.(2017)Liang, Li, and Srikant]{odin}
Shiyu Liang, Yixuan Li, and Rayadurgam Srikant.
\newblock Enhancing the reliability of out-of-distribution image detection in neural networks.
\newblock \emph{arXiv preprint arXiv:1706.02690}, 2017.

\bibitem[Liu et~al.(2020)Liu, Wang, Owens, and Li]{energy}
Weitang Liu, Xiaoyun Wang, John Owens, and Yixuan Li.
\newblock Energy-based out-of-distribution detection.
\newblock \emph{Advances in neural information processing systems}, 33:\penalty0 21464--21475, 2020.

\bibitem[Liu et~al.(2023)Liu, Lochman, and Christopher]{GEN}
Xixi Liu, Yaroslava Lochman, and Zach Christopher.
\newblock Gen: Pushing the limits of softmax-based out-of-distribution detection.
\newblock In \emph{Proceedings of the IEEE/CVF Conference on Computer Vision and Pattern Recognition}, 2023.

\bibitem[Lomonaco and Maltoni(2017)]{lomonaco2017core50}
Vincenzo Lomonaco and Davide Maltoni.
\newblock Core50: a new dataset and benchmark for continuous object recognition.
\newblock In \emph{Proceedings of the 1st Annual Conference on Robot Learning}, pages 17--26, 2017.

\bibitem[Lopez-Paz and Oquab(2016)]{lopez2016revisitingc2st}
David Lopez-Paz and Maxime Oquab.
\newblock Revisiting classifier two-sample tests.
\newblock \emph{arXiv preprint arXiv:1610.06545}, 2016.

\bibitem[Lopez-Paz and Ranzato(2017)]{gem}
David Lopez-Paz and Marc'Aurelio Ranzato.
\newblock Gradient episodic memory for continual learning.
\newblock \emph{Advances in neural information processing systems}, 30, 2017.

\bibitem[McCloskey and Cohen(1989)]{1989CF}
Michael McCloskey and Neal~J Cohen.
\newblock Catastrophic interference in connectionist networks: The sequential learning problem.
\newblock In \emph{Psychology of learning and motivation}, pages 109--165. Elsevier, 1989.

\bibitem[Miao et~al.(2024)Miao, Pang, Nguyen, Fang, Zheng, and Bai]{miao2024opencil}
Wenjun Miao, Guansong Pang, Trong-Tung Nguyen, Ruohang Fang, Jin Zheng, and Xiao Bai.
\newblock Opencil: Benchmarking out-of-distribution detection in class-incremental learning.
\newblock \emph{arXiv preprint arXiv:2407.06045}, 2024.

\bibitem[Morteza and Li(2022)]{morteza2022provable}
Peyman Morteza and Yixuan Li.
\newblock Provable guarantees for understanding out-of-distribution detection.
\newblock In \emph{Proceedings of the AAAI Conference on Artificial Intelligence}, pages 7831--7840, 2022.

\bibitem[Netzer et~al.(2011)Netzer, Wang, Coates, Bissacco, Wu, Ng, et~al.]{netzer2011svhn}
Yuval Netzer, Tao Wang, Adam Coates, Alessandro Bissacco, Baolin Wu, Andrew~Y Ng, et~al.
\newblock Reading digits in natural images with unsupervised feature learning.
\newblock In \emph{NIPS workshop on deep learning and unsupervised feature learning}, page~4. Granada, 2011.

\bibitem[Petit et~al.(2023)Petit, Popescu, Schindler, Picard, and Delezoide]{petit2023fetril}
Gr{\'e}goire Petit, Adrian Popescu, Hugo Schindler, David Picard, and Bertrand Delezoide.
\newblock Fetril: Feature translation for exemplar-free class-incremental learning.
\newblock In \emph{Proceedings of the IEEE/CVF winter conference on applications of computer vision}, pages 3911--3920, 2023.

\bibitem[Ritter et~al.(2018)Ritter, Botev, and Barber]{ritter2018online}
Hippolyt Ritter, Aleksandar Botev, and David Barber.
\newblock Online structured laplace approximations for overcoming catastrophic forgetting.
\newblock \emph{Advances in Neural Information Processing Systems}, 31, 2018.

\bibitem[Roady et~al.(2020)Roady, Hayes, Vaidya, and Kanan]{Roady_2020_Stream51}
Ryne Roady, Tyler~L. Hayes, Hitesh Vaidya, and Christopher Kanan.
\newblock Stream-51: Streaming classification and novelty detection from videos.
\newblock In \emph{The IEEE/CVF Conference on Computer Vision and Pattern Recognition (CVPR) Workshops}, 2020.

\bibitem[Rolnick et~al.(2019)Rolnick, Ahuja, Schwarz, Lillicrap, and Wayne]{rolnick2019experience}
David Rolnick, Arun Ahuja, Jonathan Schwarz, Timothy Lillicrap, and Gregory Wayne.
\newblock Experience replay for continual learning.
\newblock \emph{Advances in neural information processing systems}, 32, 2019.

\bibitem[Serra et~al.(2018)Serra, Suris, Miron, and Karatzoglou]{serra2018overcomingHAT}
Joan Serra, Didac Suris, Marius Miron, and Alexandros Karatzoglou.
\newblock Overcoming catastrophic forgetting with hard attention to the task.
\newblock In \emph{International conference on machine learning}, pages 4548--4557. PMLR, 2018.

\bibitem[Sun and Li(2022)]{sun2022dice}
Yiyou Sun and Yixuan Li.
\newblock Dice: Leveraging sparsification for out-of-distribution detection.
\newblock In \emph{European Conference on Computer Vision}, pages 691--708. Springer, 2022.

\bibitem[Sun et~al.(2021)Sun, Guo, and Li]{sun2021react}
Yiyou Sun, Chuan Guo, and Yixuan Li.
\newblock React: Out-of-distribution detection with rectified activations.
\newblock \emph{Advances in Neural Information Processing Systems}, 34:\penalty0 144--157, 2021.

\bibitem[Tieppo et~al.(2022)Tieppo, Santos, Barddal, and Nievola]{tieppo2022hierarchical}
Eduardo Tieppo, Roger Robson~dos Santos, Jean~Paul Barddal, and J{\'u}lio~Cesar Nievola.
\newblock Hierarchical classification of data streams: a systematic literature review.
\newblock \emph{Artificial Intelligence Review}, 55\penalty0 (4):\penalty0 3243--3282, 2022.

\bibitem[Van~de Ven et~al.(2022)Van~de Ven, Tuytelaars, and Tolias]{type_til}
Gido~M Van~de Ven, Tinne Tuytelaars, and Andreas~S Tolias.
\newblock Three types of incremental learning.
\newblock \emph{Nature Machine Intelligence}, 4\penalty0 (12):\penalty0 1185--1197, 2022.

\bibitem[Vinyals et~al.(2016)Vinyals, Blundell, Lillicrap, Wierstra, et~al.]{vinyals2016minii}
Oriol Vinyals, Charles Blundell, Timothy Lillicrap, Daan Wierstra, et~al.
\newblock Matching networks for one shot learning.
\newblock \emph{Advances in neural information processing systems}, 29, 2016.

\bibitem[Wang et~al.(2024)Wang, Zhang, Su, and Zhu]{wang2024comprehensive}
Liyuan Wang, Xingxing Zhang, Hang Su, and Jun Zhu.
\newblock A comprehensive survey of continual learning: theory, method and application.
\newblock \emph{IEEE Transactions on Pattern Analysis and Machine Intelligence}, 2024.

\bibitem[Yang et~al.(2024)Yang, Zhou, Li, and Liu]{yang2024generalizedood}
Jingkang Yang, Kaiyang Zhou, Yixuan Li, and Ziwei Liu.
\newblock Generalized out-of-distribution detection: A survey.
\newblock \emph{International Journal of Computer Vision}, pages 1--28, 2024.

\bibitem[Yu et~al.(2024)Yu, Rosing, and Guo]{yu2024evolve}
Xiaofan Yu, Tajana Rosing, and Yunhui Guo.
\newblock Evolve: Enhancing unsupervised continual learning with multiple experts.
\newblock In \emph{Proceedings of the IEEE/CVF winter conference on applications of computer vision}, pages 2366--2377, 2024.

\bibitem[Yu et~al.(2023)Yu, Shin, Lee, Jun, and Lee]{feature_norm}
Yeonguk Yu, Sungho Shin, Seongju Lee, Changhyun Jun, and Kyoobin Lee.
\newblock Block selection method for using feature norm in out-of-distribution detection.
\newblock In \emph{Proceedings of the IEEE/CVF Conference on Computer Vision and Pattern Recognition}, pages 15701--15711, 2023.

\bibitem[Yuan et~al.(2023)Yuan, He, Han, and Yin]{yuan2023lhact}
Yue Yuan, Rundong He, Zhongyi Han, and Yilong Yin.
\newblock Lhact: Rectifying extremely low and high activations for out-of-distribution detection.
\newblock In \emph{Proceedings of the 31st ACM International Conference on Multimedia}, pages 8105--8113, 2023.

\bibitem[Zenke et~al.(2017)Zenke, Poole, and Ganguli]{zenke2017continualSI}
Friedemann Zenke, Ben Poole, and Surya Ganguli.
\newblock Continual learning through synaptic intelligence.
\newblock In \emph{International conference on machine learning}, pages 3987--3995. PMLR, 2017.

\end{thebibliography}
}
\clearpage
\setcounter{page}{1}
\maketitlesupplementary
\section{Supplementary}
In this supplementary, we include more details on the following aspects:
\begin{itemize}
    \item We report experimental results and conduct analysis of single C2ST performance in Section \ref{sup_sec:sing c2st}.
    \item We provide detailed experimental results of C2ST in Section \ref{sup_sec:c2st}.
    \item We present visualizations of performance sensitivity to depth in Section \ref{sup_sec:depth}.
    \item We provide detailed results of different memory sizes in Section \ref{sup_sec:memory}.
    \item We discuss the source-target classifier architecture in  Section \ref{sup_sec:archi}.
\end{itemize}

\subsection{Experimental Results and Analysis of single C2ST}
\label{sup_sec:sing c2st}
In Section \ref{sub_sec:2st}, we critically discussed the limitation of the sequential covariate shift detection method \cite{jang2022sequential}. Their methodology presents a fundamental trade-off dilemma: While a unified classifier architecture demonstrates computational efficiency, its binary discriminative framework intrinsically lacks the capacity for task-id recognition. Conversely, maintaining specific classifiers per task effectively preserves discriminability inter tasks at the cost of increased resource overhead with increasing tasks. In the main text, we conducted a theoretical analysis of the second method and presented experimental results. Now, we turn to the first method, namely single-C2ST, for discussion.

\begin{figure}[ht]
  \centering
   \includegraphics[width=\linewidth]{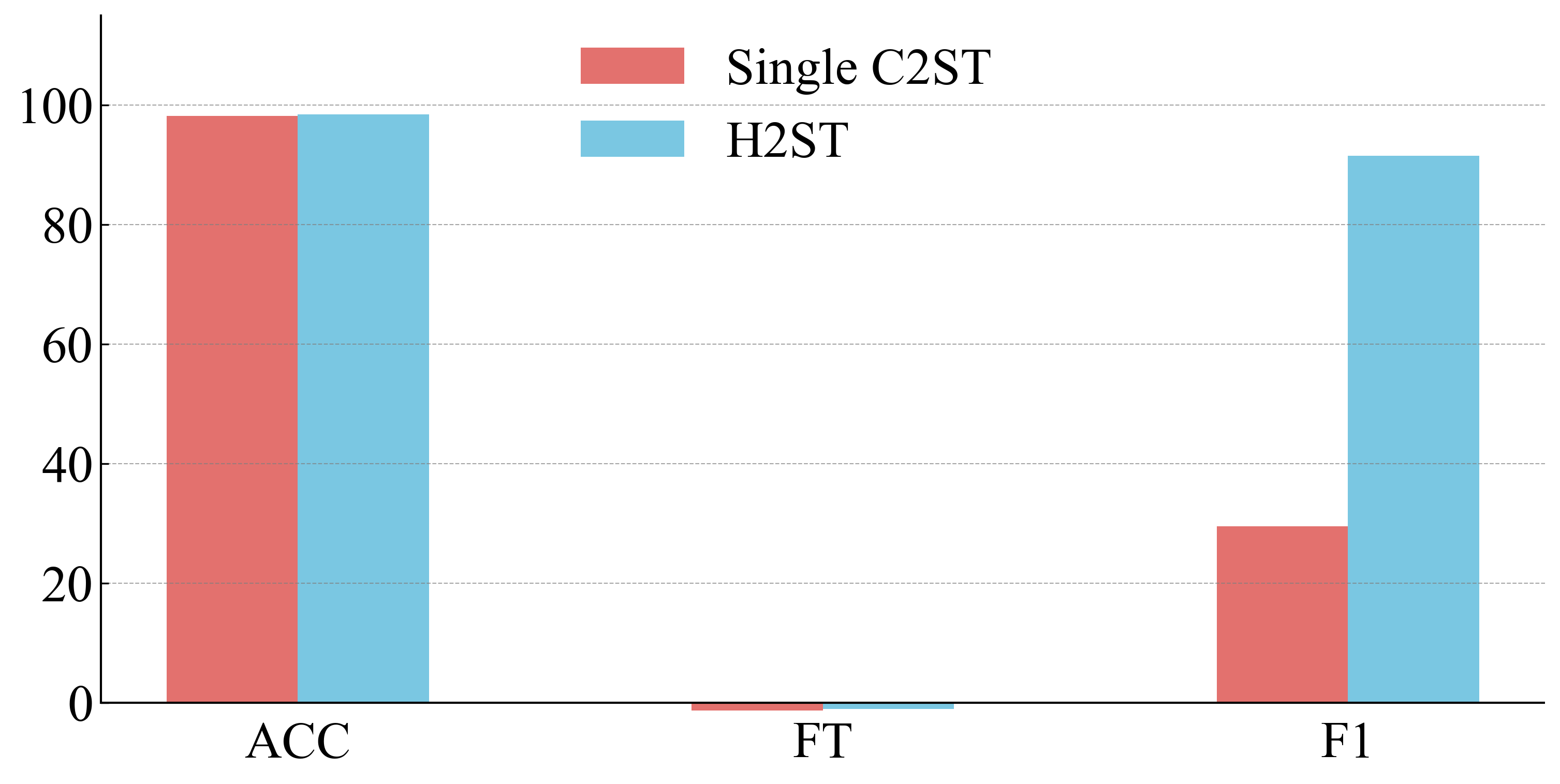}
   \caption{TIL and OOD detection performance of single C2ST and H2ST. H2ST demonstrates superior OOD detection performance.}
   \label{fig:singlec2st}
\end{figure}

In the single-C2ST, a unified source-target classifier is maintained. For the classifier, all previously learned tasks are treated as ID, so the source samples will be drawn as evenly as possible from the respective memory buffers of all learned tasks. This architecture enables each sample to undergo only a single evaluation pass through the classifier, thereby achieving significant computational resource efficiency. However, this method exhibits two significant limitations. Firstly, as previously discussed, it inherently lacks the capability for task-id prediction, making it unsuitable for open-world TIL. Secondly, as incremental learning progresses, the number of ID tasks increases substantially and the number of source samples drawn from each task decreases, resulting in degraded classifier performance. The detailed OOD and TIL detection performance are respectively presented in Table \ref{tab:single_ood} and Table \ref{tab:single_til}. The task-id in the tables refers to the identity of the most recently seen task. TP, FP, TN, and FN represent the IDs correctly predicted, OODs incorrectly predicted as IDs, OODs correctly predicted, and IDs incorrectly predicted as OODs, respectively. As shown in Table \ref{tab:single_ood}, almost all OOD samples can be detected, but the majority of ID samples will also be incorrectly identified as OOD. When the number of learned tasks increases, the composition of the source sample becomes more complex, and the ID samples only belong to one of the tasks, so it becomes more difficult to correctly predict the ID test sample as ID. The OOD target sample does not match any of them, so the classifier can still accurately identify them as OOD, providing sufficient training data for TIL. As depicted in Fig. \ref{fig:singlec2st}, H2ST and single-C2ST exhibit comparable performance in TIL. However, the hierarchical architecture demonstrates considerable improvement in OOD detection, with an average increase of $62.06\%$ in F1.
\begin{table}[t]
  \centering
  \small
    \begin{tabular}{c|ccccc}
    \hline
    \multirow{2}[2]{*}{Task-id} & \multicolumn{4}{c}{OOD Detection Result} & \multirow{2}[2]{*}{F1 Score} \\
          & TP    & FP    & TN    & FN    &  \\
    \hline
    1     & 1920 & 30 & 5970 & 80 & 97.22  \\
    2     & 170 & 0 & 6000 & 3830 & 8.15  \\
    3     & 150 & 0 & 6000 & 3850 & 7.23  \\
    4     & 110 & 0 & 6000 & 3890 & 5.35  \\
    Average &       &       &       &       & 29.49  \\
    \hline
    \end{tabular}%
    \caption{Detailed OOD detection performance of single-C2ST on MNIST dataset with GEM.}
  \label{tab:single_ood}%
\end{table}%

\begin{table}[t]
  \centering
  \small
    \begin{tabular}{c|ccccc}
    \hline
    \multirow{2}[2]{*}{Task-id} & \multicolumn{5}{c}{Task Incremental Learning Accuracy} \\
          & Task1 & Task2 & Task3 & Task4 & Task5 \\
    \hline
    1     & 99.49  & 58.27  & 54.70  & 52.89  & 51.28  \\
    2     & 99.35  & 99.08  & 61.68  & 70.71  & 32.14  \\
    3     & 98.43  & 98.49  & 98.44  & 54.65  & 43.65  \\
    4     & 98.75  & 97.95  & 97.44  & 99.36  & 46.62  \\
    5     & 98.84  & 97.46  & 96.84  & 98.24  & 99.49  \\
\cline{2-6}    Average & ACC:  & 98.17  &       & FT:   & -1.25  \\
    \hline
    \end{tabular}%
    \caption{Detailed TIL performance of single-C2ST on MNIST dataset with GEM.}
  \label{tab:single_til}%
\end{table}%

\subsection{Detailed Results of C2ST}
\label{sup_sec:c2st}
We present a comparative analysis between H2ST and C2ST in Figure \ref{fig:parallel}. Here we present comprehensive C2ST results, detailed in Table \ref{tab:sup_c2st_ood} and Table \ref{tab:sup_c2st_til}. While C2ST exhibits improved OOD detection performance compared to baseline methods, it still demonstrates a measurable performance gap relative to our proposed H2ST.
\subsection{Visualization of Performance Sensitivity to Depth}
\label{sup_sec:depth}
\begin{figure*}
  \centering
  \begin{subfigure}{0.49\linewidth}
    \includegraphics[width=\linewidth]{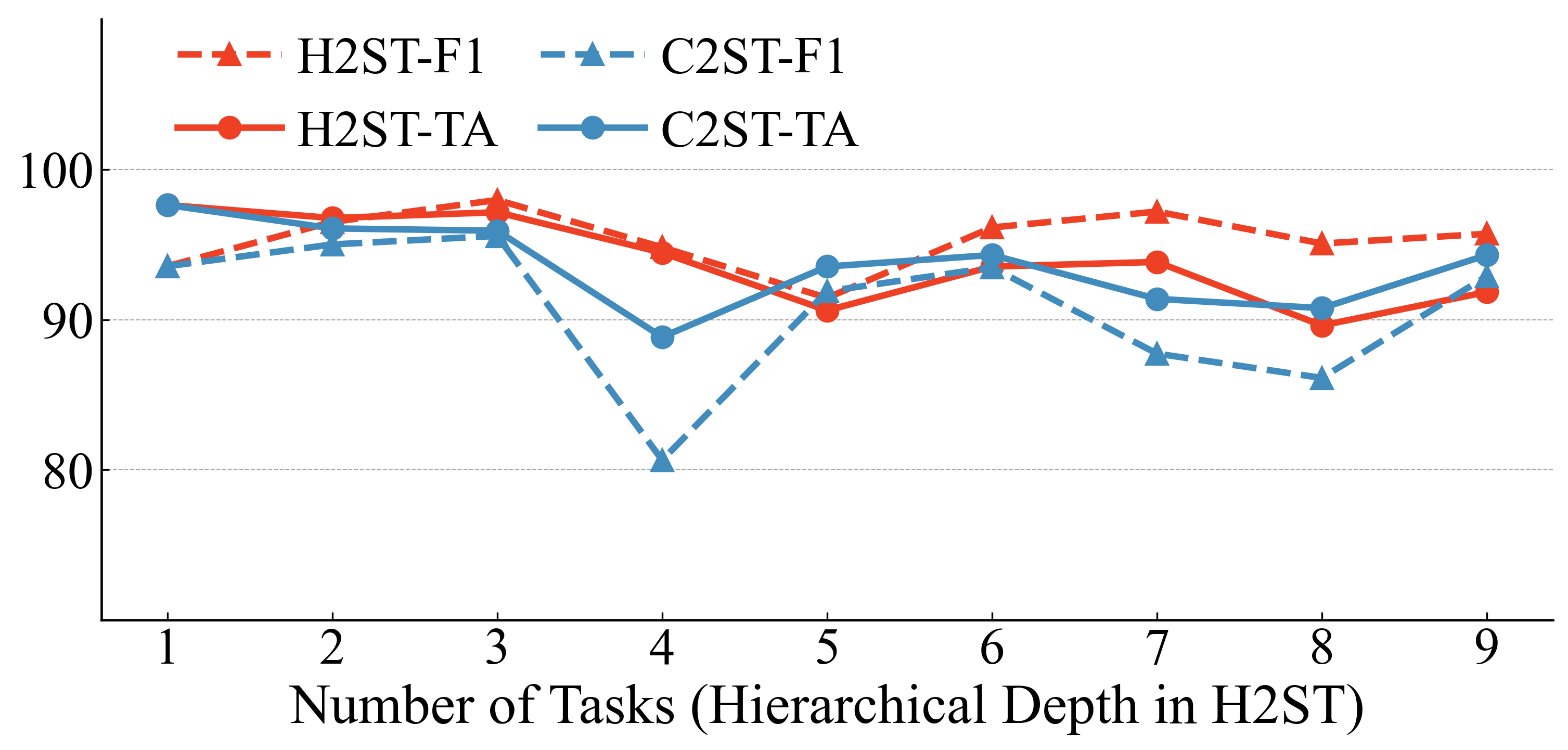}
    \caption{CoRe50 dataset with GEM.}
    \label{fig:core_gem}
  \end{subfigure}
  \hfill
  \begin{subfigure}{0.49\linewidth}
    \includegraphics[width=\linewidth]{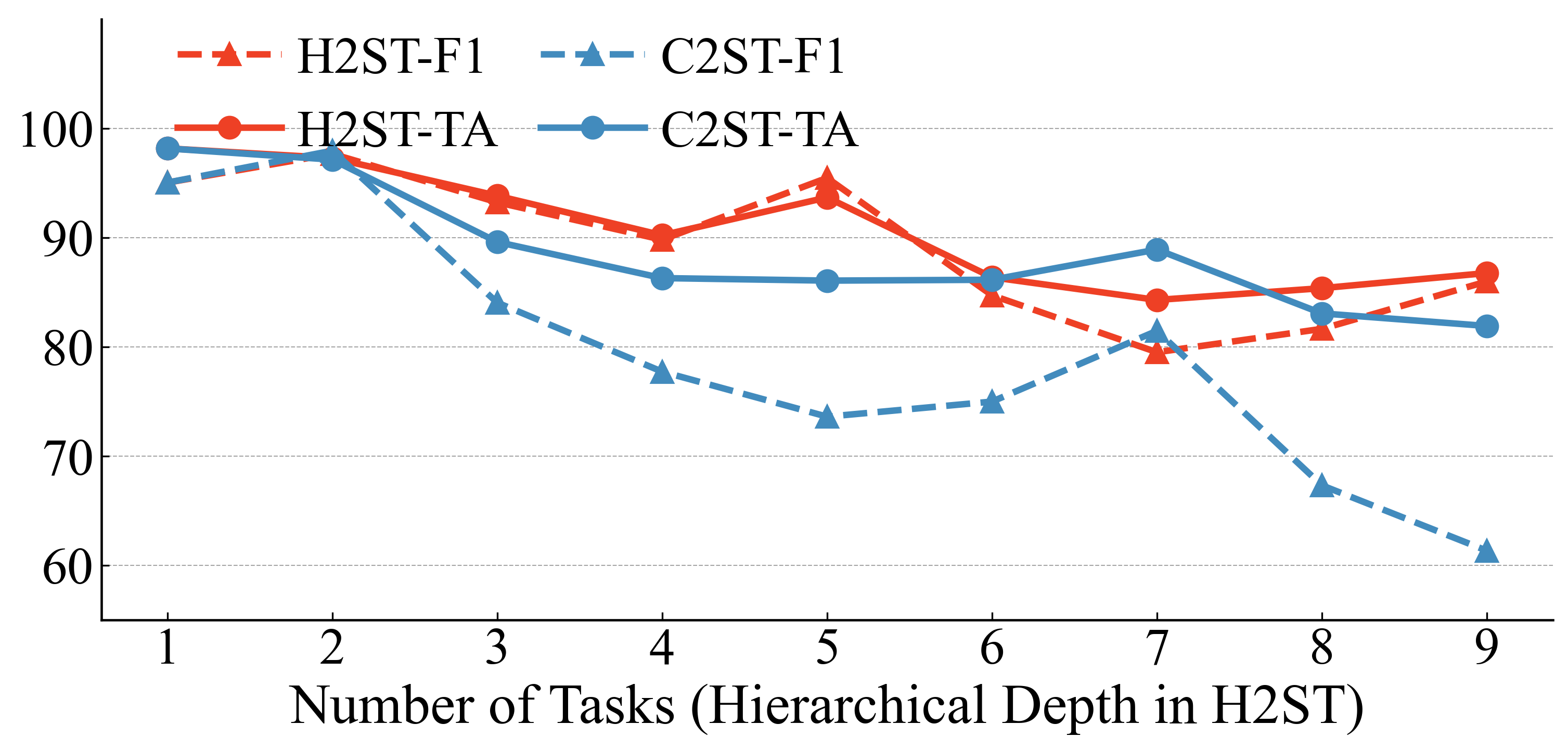}
    \caption{Stream-51 dataset with ER.}
    \label{fig:stream_er}
  \end{subfigure}
  \\
  \begin{subfigure}{0.49\linewidth}
    \includegraphics[width=\linewidth]{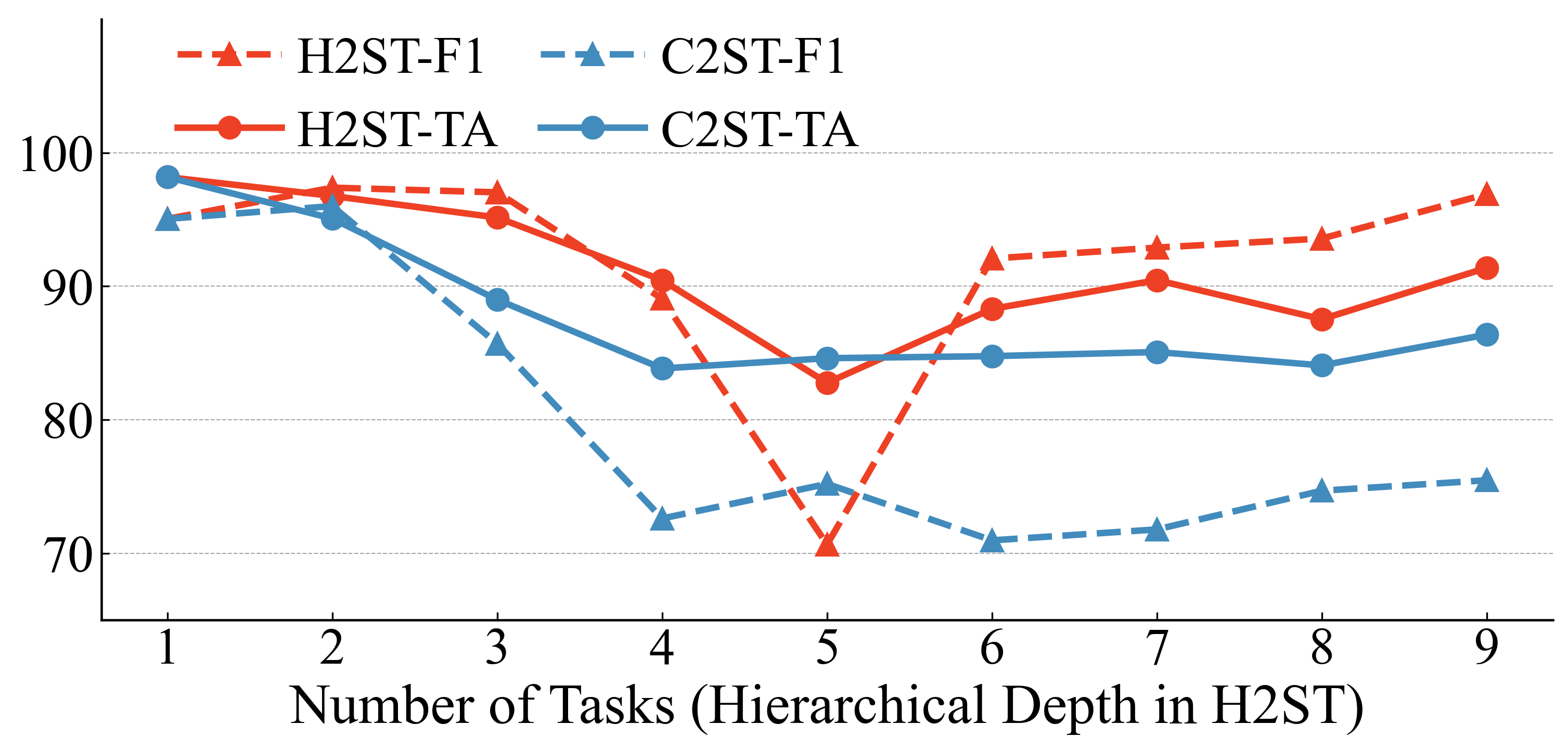}
    \caption{Stream-51 dataset with GEM.}
    \label{fig:stream_gem}
  \end{subfigure}  
  \hfill
  \begin{subfigure}{0.49\linewidth}
    \includegraphics[width=\linewidth]{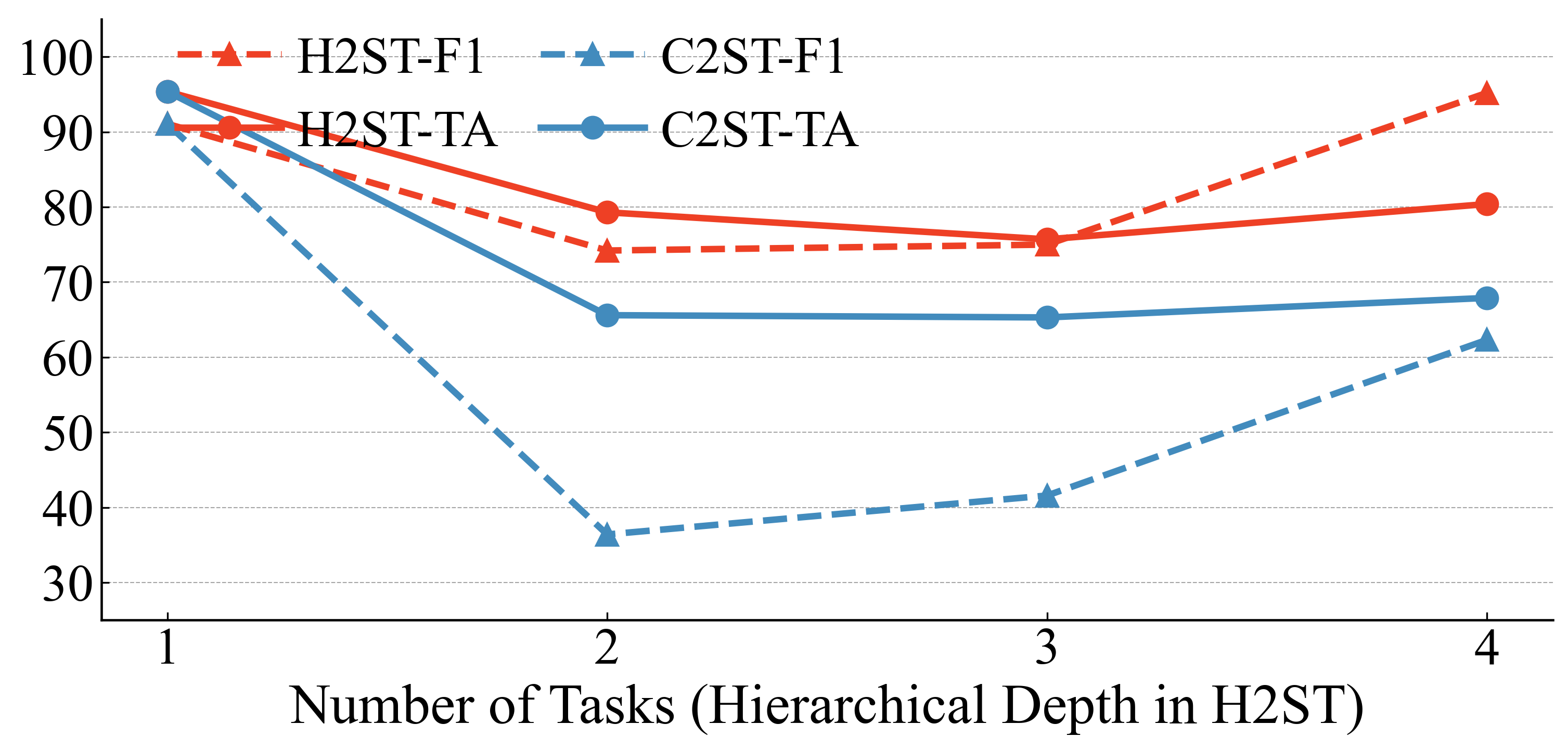}
    \caption{Mini-ImageNet dataset with GEM.}
    \label{fig:mini_gem}
    \end{subfigure}   
    \\
  \begin{subfigure}{0.49\linewidth}
    \includegraphics[width=\linewidth]{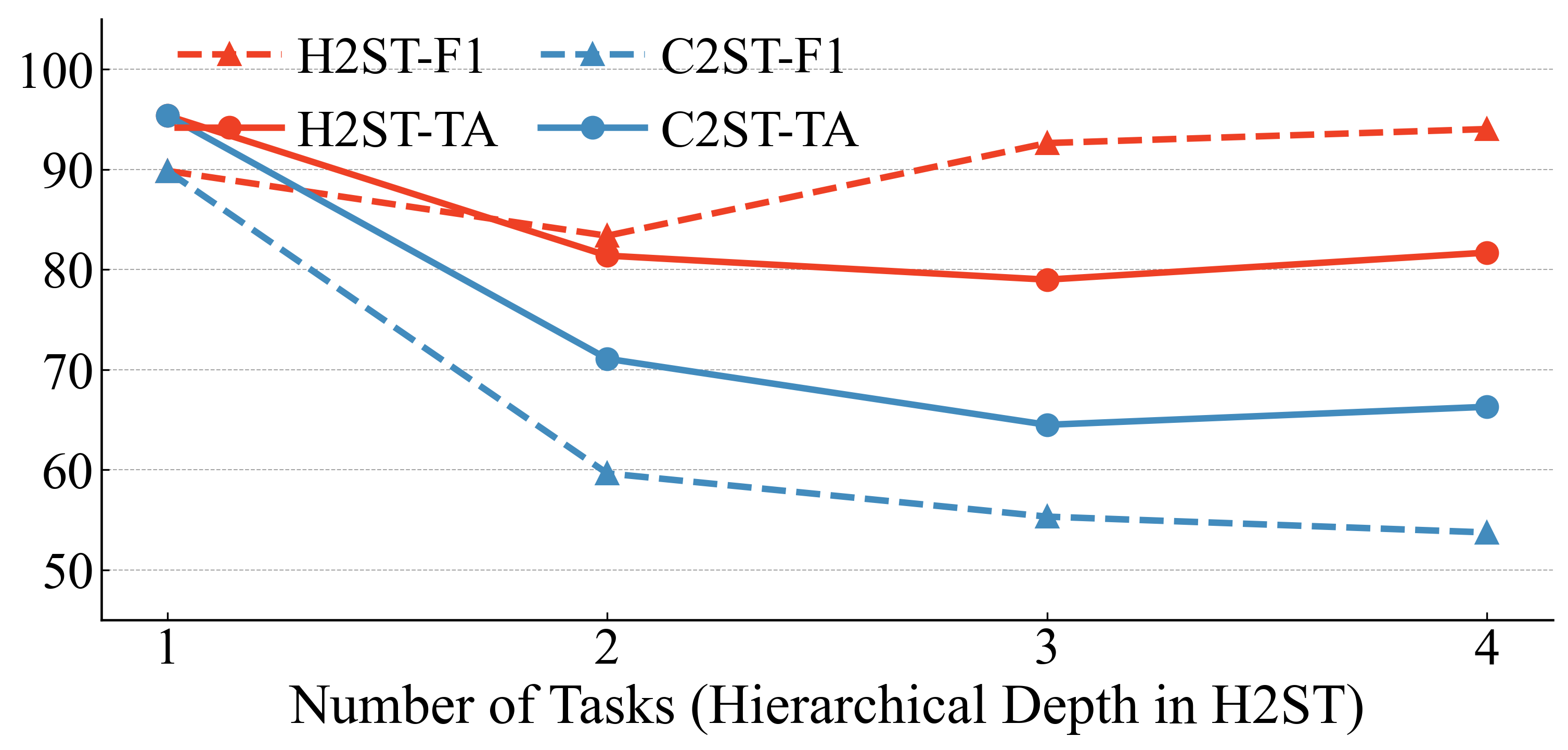}
    \caption{CIFAR-100 dataset with GEM.}
    \label{fig:c100_gem}
  \end{subfigure}  
   \hfill
   \begin{subfigure}{0.49\linewidth}
    \includegraphics[width=\linewidth]{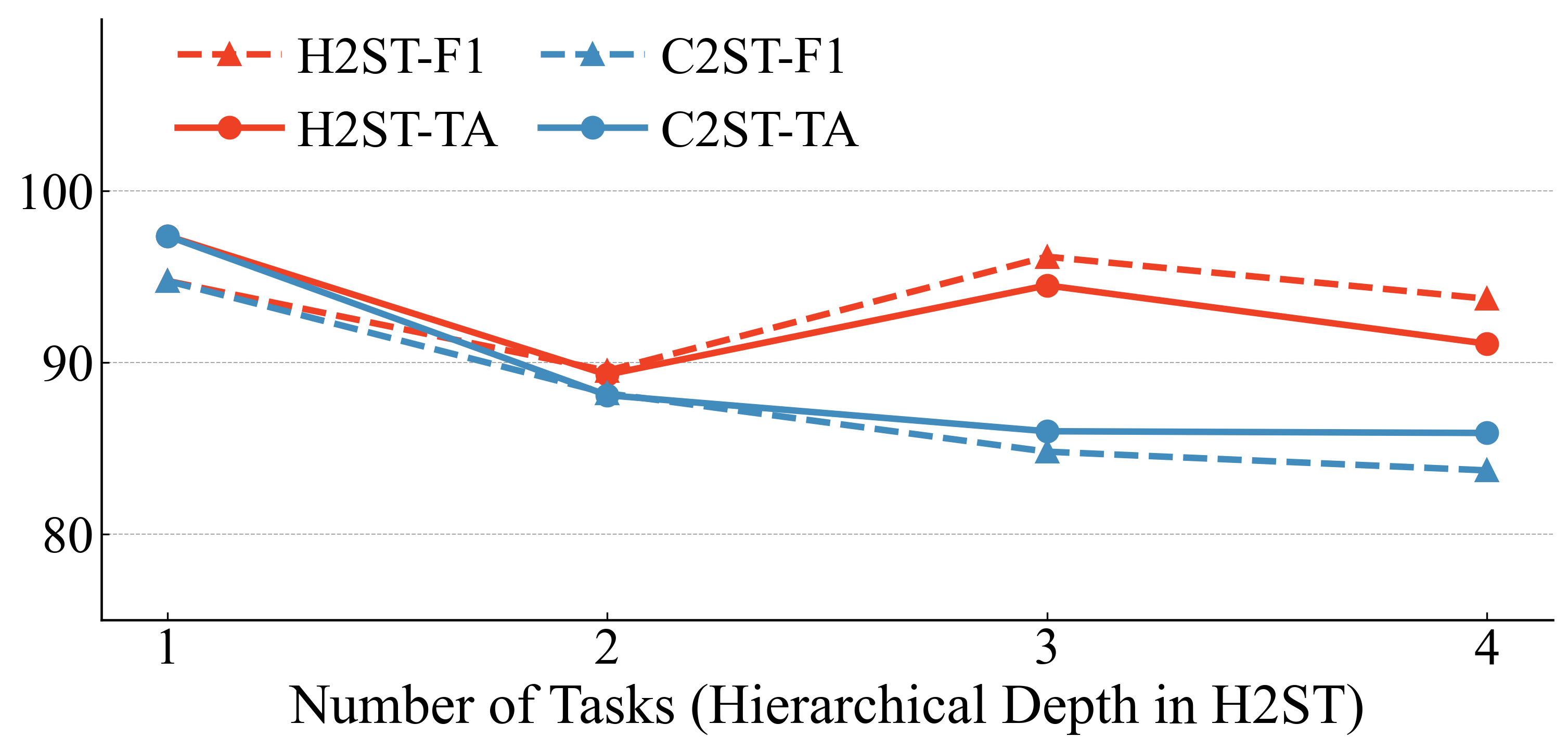}
    \caption{CIFAR-10 dataset with GEM.}
    \label{fig:c10_gem}
  \end{subfigure}  
  \caption{Performance sensitivity to depth of (a) CoRe50 dataset with GEM, (b)Stream-51 dataset with ER, (c)Stream-51 dataset with GEM, (d)Mini-ImageNet dataset with GEM, (e)CIFAR-100 dataset with GEM and (f) CIFAR-10 dataset with GEM.}
  \label{fig:supplx_depth}
\end{figure*}

While Fig. \ref{fig:depth} visualizes the depth sensitivity analysis for CoRe50 with ER, we extend this analysis to other cases in Fig. \ref{fig:supplx_depth}. The results demonstrate that H2ST achieves superior overall performance and shows greater stability. Particularly in cases with a large number of classes per task, such as CIFAR-100 and Mini-ImageNet, the performance of C2ST deteriorates rapidly as the number of learned tasks increases. In contrast, H2ST consistently maintains stable performance throughout the incremental learning process.
\subsection{Detailed Results of Different Memory Sizes}
\label{sup_sec:memory}
In Fig. \ref{fig:memory}, we present the trend of various metrics relative to memory size per task. We now provide detailed results of different memory sizes in Table \ref{tab:sup_memory_ood} and Table \ref{tab:sup_memory_til}. Memory size significantly impacts OOD detection, and we conduct an in-depth analysis. In H2ST, source samples are randomly drawn from memory buffers, with the fundamental assumption that these samples sufficiently represent the task distribution. An insufficient memory size leads to insufficient coverage of the task distribution, inadequate sample diversity, and increased distribution estimation bias. Conversely, simply increasing memory size is not an optimal solution, as the information gain from additional samples becomes negligible beyond a certain extent and larger memory directly leads to higher overhead. Therefore, finding an optimal memory size that balances representativeness and computational efficiency is crucial for effective OOD detection.
\subsection{Different Source-Target Classifier Architectures}
\label{sup_sec:archi}
\begin{figure}[t]
  \centering
   \includegraphics[width=\linewidth]{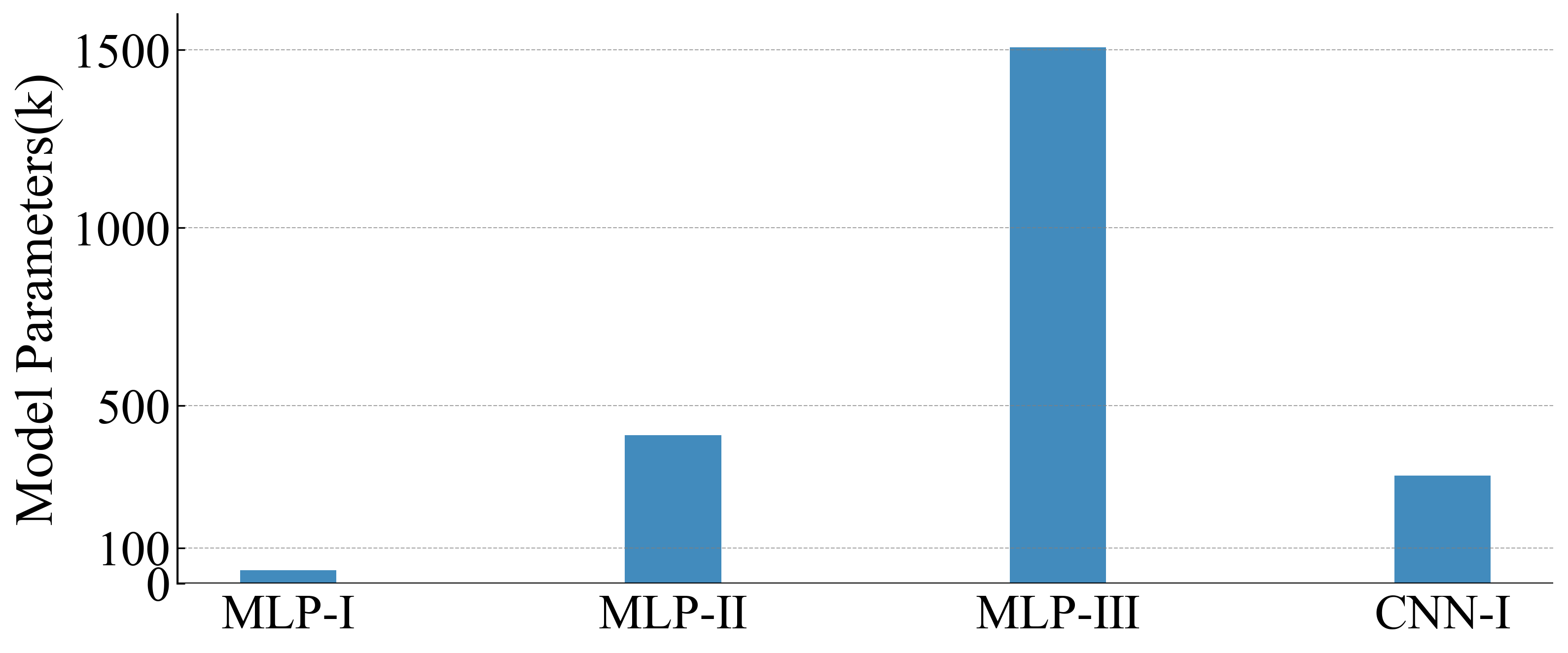}
   \caption{Parameters of different source-target  classifier architectures.}
   \label{fig:para}
\end{figure}

\begin{figure}[t]
  \centering
   \includegraphics[width=\linewidth]{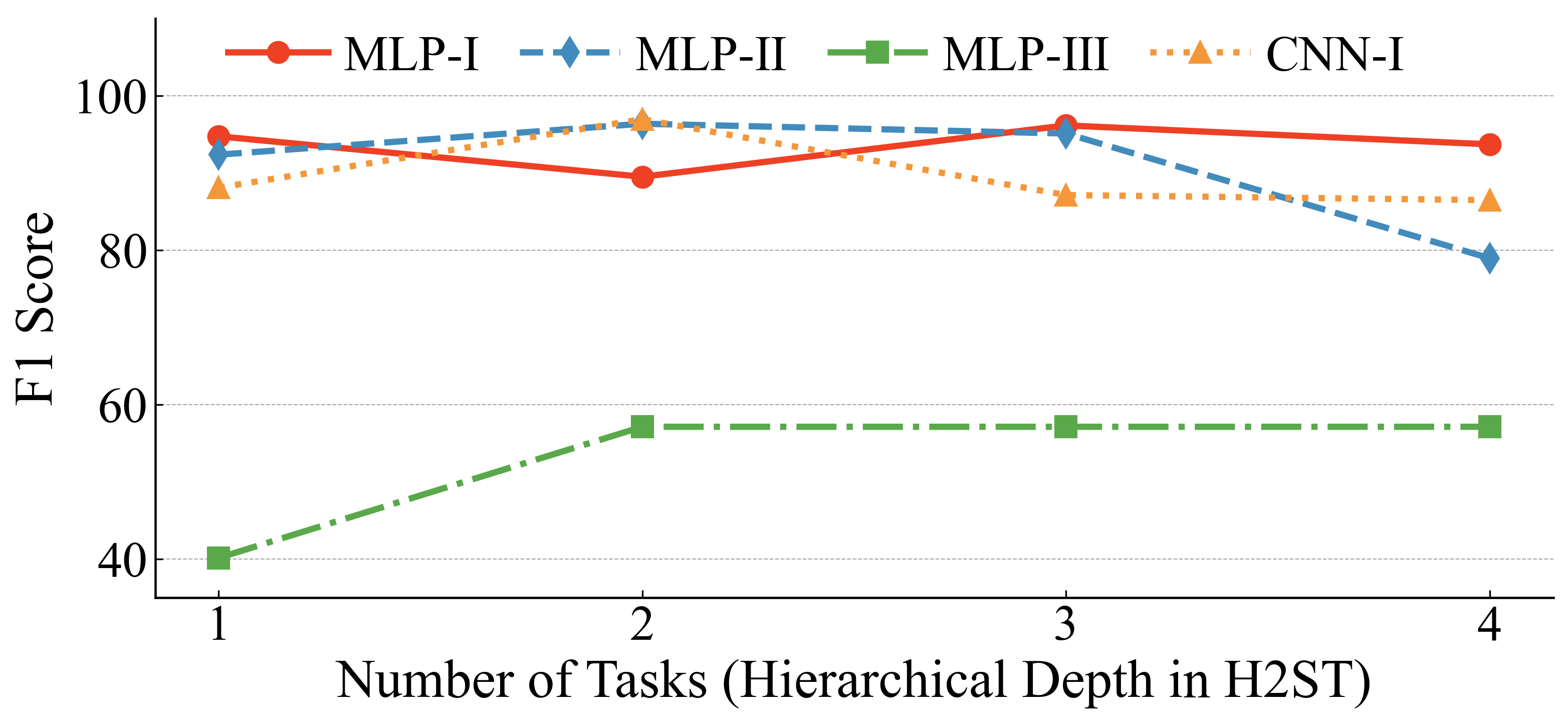}
   \caption{F1 score sensitivity to depth of different source-target classifier architectures.}
   \label{fig:sensi_archi}
\end{figure}

In Section \ref{sub_sec:exp_setup}, we employ a fully connected neural network with a single hidden layer of 128 ReLU units, denoted as MLP-\uppercase\expandafter{\romannumeral1}, as the source-target classifier. While this lightweight architecture demonstrates promising performance, we further explore alternative architectures. Specifically, we investigate deeper fully connected neural network with five hidden layers (MLP-\uppercase\expandafter{\romannumeral2}) and ten hidden layers (MLP-\uppercase\expandafter{\romannumeral3}), along with convolutional neural network with four convolutional layers (CNN-\uppercase\expandafter{\romannumeral1}), to provide comprehensive architectural comparisons. Fig. \ref{fig:para} illustrates the number of parameters of these models. We conduct experiments on CIFAR-10, with the F1 scores shown in Fig. \ref{fig:sensi_archi} and the average metrics summarized in Table \ref{tab:sup_archi}. The MLP-\uppercase\expandafter{\romannumeral1} with the fewest parameters achieves the best OOD detection effect, while the MLP-\uppercase\expandafter{\romannumeral3} with the most parameters performs the worst, misclassifying a large number of OOD samples as ID. This phenomenon is particularly relevant in continual learning characterized by non-stationary data streams, where source-target classifiers must be updated online to adapt to new distributions. While deeper models might theoretically offer greater representational capacity, their increased complexity hinders rapid adaptation to new distributions. In contrast, models with simpler architecture demonstrate superior adaptability, enabling faster adjustments in response to distributional shifts. Moreover, since each classifier only performs binary classification, simpler models are generally sufficient to meet the demands. Furthermore, their lower computational overhead makes them more practical for applications.

\begin{table}[t]
  \centering
  %\small
  %\renewcommand{\arraystretch}{1.20}
    \begin{tabular}{ccccc}
    \hline
    Architecture    & ACC$\uparrow$   & FT$\uparrow$    & F1$\uparrow$    & TA$\uparrow$ \\
    \hline
    MLP-\uppercase\expandafter{\romannumeral1} & \textbf{84.71}  & \textbf{-7.90} & \textbf{93.54}  & \textbf{92.82}   \\
    MLP-\uppercase\expandafter{\romannumeral2} & 83.39  & -6.83  & 90.72  &  78.67 \\
    MLP-\uppercase\expandafter{\romannumeral3} & 51.14 & -7.05  & 52.90  &  11.38 \\
    CNN-\uppercase\expandafter{\romannumeral1} & 84.37  & -6.83  & 89.68 &  79.16 \\
    \hline
    \end{tabular}
    \caption{Performance comparison across different source-target classifier architectures.} 
  \label{tab:sup_archi}
\end{table}

\begin{table*}[ht]
\small
\renewcommand{\arraystretch}{1.25}
%\scriptsize
  \centering
  \resizebox{\linewidth}{!}{
    \begin{tabular}{c|cc|cc|cc|cc|cc|cc|cc|cc}
    \hline
    &    \multicolumn{14}{c}{Dataset}    &       &  \\
\cline{2-15}    TIL & \multicolumn{2}{c}{MNIST} & \multicolumn{2}{c}{SVHN} & \multicolumn{2}{c}{CIFAR-10} & \multicolumn{2}{c}{CIFAR-100} & \multicolumn{2}{c}{Mini-ImageNet} & \multicolumn{2}{c}{CoRe50} & \multicolumn{2}{c}{Stream-51} & \multicolumn{2}{c}{Average} \\
\cline{2-17}   & F1$\uparrow$    & TA$\uparrow$ & F1$\uparrow$    & TA$\uparrow$      & F1$\uparrow$    &TA$\uparrow$       & F1$\uparrow$    &TA$\uparrow$       & F1$\uparrow$    & TA$\uparrow$      & F1$\uparrow$   & TA$\uparrow$       & F1$\uparrow$   & TA$\uparrow$       & F1$\uparrow$   & TA$\uparrow$\\
    \hline
    \multirow{1}{*}{ER \cite{rolnick2019experience}} & 85.46  & 90.01  & 74.13  & 83.05  & 86.04  & 88.72  & 60.18  & 73.09  & 56.67  & 73.57  & 86.15  & 91.50  & 79.29  & 88.60  & 75.42  & 84.08  \\
    \multirow{1}{*}{GEM \cite{gem}} &   86.34  & 90.31  & 74.92  & 83.65  & 87.88  & 89.35  & 64.65  & 74.32  & 57.84  & 73.54  & 90.77  & 93.64  & 79.72  & 87.89  & 77.45  & 84.67  \\
    \hline
    \end{tabular}
    }
  \caption{OOD detection performance of C2ST.}
  \label{tab:sup_c2st_ood}
\end{table*}

\begin{table*}[ht]
\small
\renewcommand{\arraystretch}{1.25}
%\scriptsize
  \centering
  \resizebox{\linewidth}{!}{
    \begin{tabular}{c|cc|cc|cc|cc|cc|cc|cc|cc}
    \hline
    &    \multicolumn{14}{c}{Dataset}    &       &  \\
\cline{2-15}    TIL & \multicolumn{2}{c}{MNIST} & \multicolumn{2}{c}{SVHN} & \multicolumn{2}{c}{CIFAR-10} & \multicolumn{2}{c}{CIFAR-100} & \multicolumn{2}{c}{Mini-ImageNet} & \multicolumn{2}{c}{CoRe50} & \multicolumn{2}{c}{Stream-51} & \multicolumn{2}{c}{Average} \\
\cline{2-17}   & ACC$\uparrow$    & FT$\uparrow$ & ACC$\uparrow$    & FT$\uparrow$      & ACC$\uparrow$    &FT$\uparrow$       & ACC$\uparrow$    &FT$\uparrow$       & ACC$\uparrow$    & FT$\uparrow$      & ACC$\uparrow$   & FT$\uparrow$       & ACC$\uparrow$   & FT$\uparrow$       & ACC$\uparrow$   & FT$\uparrow$\\
    \hline
    \multirow{1}{*}{ER \cite{rolnick2019experience}} & 96.14  & -1.73  & 94.23  & -3.16  & 83.72  & -9.01  & 45.56  & -14.06  & 31.02  & -11.76  & 75.04  & -4.70  & 67.56  & -8.66  & 70.47  & -7.58  \\
    \multirow{1}{*}{GEM \cite{gem}} &   98.67  & -0.56  & 92.64  & -5.20  & 84.41  & -7.80  & 44.97  & -14.40  & 32.15  & -10.86  & 77.03  & -3.50  & 67.22  & -11.68  & 71.01  & -7.71  \\
    \hline
    \end{tabular}
    }
  \caption{TIL performance of C2ST.}
  \label{tab:sup_c2st_til}
\end{table*}

\begin{table*}[ht]
\small
\renewcommand{\arraystretch}{1.25}
%\scriptsize
  \centering
  \resizebox{\linewidth}{!}{
    \begin{tabular}{c|c|cc|cc|cc|cc|cc|cc|cc|cc}
    \hline
    &       & \multicolumn{14}{c}{Dataset}    &       &  \\
\cline{3-16}    TIL & Memory Size & \multicolumn{2}{c}{MNIST} & \multicolumn{2}{c}{SVHN} & \multicolumn{2}{c}{CIFAR-10} & \multicolumn{2}{c}{CIFAR-100} & \multicolumn{2}{c}{Mini-ImageNet} & \multicolumn{2}{c}{CoRe50} & \multicolumn{2}{c}{Stream-51} & \multicolumn{2}{c}{Average} \\
\cline{3-18}          &       & F1$\uparrow$    & TA$\uparrow$ & F1$\uparrow$    & TA$\uparrow$      & F1$\uparrow$    &TA$\uparrow$       & F1$\uparrow$    &TA$\uparrow$       & F1$\uparrow$    & TA$\uparrow$      & F1$\uparrow$   & TA$\uparrow$       & F1$\uparrow$   & TA$\uparrow$       & F1$\uparrow$   & TA$\uparrow$\\
    \hline
    \multirow{4}{*}{ER \cite{rolnick2019experience}}& 40  & 23.82  & 68.16  & 63.65  & 78.27  & 52.89  & 73.78  & 31.31  & 66.69  & 31.60  & 67.23  & 44.68  & 76.09  & 39.11  & 73.77  & 41.01  & 72.00  \\
    & 100 &  64.77  & 80.14  & 70.20  & 81.86  & 84.65  & 87.09  & 56.20  & 71.81  & 55.60  & 72.88  & 83.90  & 88.69  & 66.78  & 81.57  & 68.87  & 80.58  \\
    & 200  &  92.03  & 93.78  & 77.60  & 84.60  & 88.89  & 89.59  & 84.21  & 82.02  & 79.34  & 81.59  & 94.11  & 94.06  & 89.24  & 90.68  & 86.49  & 88.05  \\
    & 300 &  95.21  & 96.06  & 82.05  & 85.71  & 93.56  & 92.43  & 92.18  & 86.76  & 85.37  & 82.16  & 97.47  & 95.23  & 94.27  & 92.85  & 91.44  & 90.17  \\

    \hline
    \multirow{4}{*}{GEM \cite{gem}}& 40  & 23.53  & 68.03  & 69.45  & 79.92  & 57.41  & 75.13  & 34.03  & 66.42  & 30.89  & 66.78  & 52.18  & 77.67  & 41.05  & 74.39  & 44.08  & 72.62  \\
    & 100 &68.86  & 82.29  & 75.83  & 83.66  & 78.57  & 84.26  & 55.54  & 72.24  & 54.20  & 72.46  & 87.62  & 90.09  & 72.82  & 83.34  & 70.49  & 81.19  \\
    & 200  &  91.55  & 93.43  & 77.87  & 85.20  & 93.54  & 92.82  & 89.98  & 84.37  & 83.88  & 82.69  & 95.38  & 93.94  & 91.63  & 91.23  & 89.12  & 89.10  \\
    & 300 & 95.41  & 96.09  & 77.63  & 83.73  & 94.41  & 93.01  & 94.23  & 85.16  & 92.90  & 85.53  & 98.01  & 95.69  & 96.02  & 94.23  & 92.66  & 90.49  \\
    \hline
    \end{tabular}
    }
  \caption{Comparison of the OOD detection performance across different memory sizes.}
  \label{tab:sup_memory_ood}
\end{table*}

\begin{table*}[ht]
\small
\renewcommand{\arraystretch}{1.25}
%\scriptsize
  \centering
  \resizebox{\linewidth}{!}{
    \begin{tabular}{c|c|cc|cc|cc|cc|cc|cc|cc|cc}
    \hline
    &       & \multicolumn{14}{c}{Dataset}    &       &  \\
\cline{3-16}    TIL & Memory Size & \multicolumn{2}{c}{MNIST} & \multicolumn{2}{c}{SVHN} & \multicolumn{2}{c}{CIFAR-10} & \multicolumn{2}{c}{CIFAR-100} & \multicolumn{2}{c}{Mini-ImageNet} & \multicolumn{2}{c}{CoRe50} & \multicolumn{2}{c}{Stream-51} & \multicolumn{2}{c}{Average} \\
\cline{3-18} &  & ACC$\uparrow$    & FT$\uparrow$ & ACC$\uparrow$    & FT$\uparrow$      & ACC$\uparrow$    &FT$\uparrow$       & ACC$\uparrow$    &FT$\uparrow$       & ACC$\uparrow$    & FT$\uparrow$      & ACC$\uparrow$   & FT$\uparrow$       & ACC$\uparrow$   & FT$\uparrow$       & ACC$\uparrow$   & FT$\uparrow$\\
    \hline

    \multirow{4}{*}{ER \cite{rolnick2019experience}}& 40  & 97.34  & -2.21  & 91.95  & -6.34  & 79.50  & -14.15  & 41.12  & -19.14  & 28.06  & -16.79  & 72.56  & -10.74  & 64.57  & -11.23  & 67.87  & -11.51  \\
    & 100 & 97.80  & -1.45  & 92.62  & -5.67  & 79.30  & -13.80  & 43.68  & -15.70  & 29.99  & -15.09  & 77.13  & -4.31  & 74.37  & -8.19  & 70.70  & -9.17  \\
    & 200  & 98.49 & -0.80  & 93.45  & -4.24  & 84.34  & -8.98  & 45.09  & -14.23  & 31.91  & -10.56  & 78.26  & -1.42  & 74.33  & -5.50  & 72.27  & -6.53  \\
    & 300 &  98.71  & -0.32  & 94.94  & -2.59  & 85.76  & -6.76  & 45.70  & -12.45  & 33.27  & -9.81  & 79.07  & -2.93  & 75.22  & 1.98  & 73.24  & -4.70  \\

    \hline
    \multirow{4}{*}{GEM \cite{gem}}& 40  & 97.37  & -2.30  & 93.32  & -4.76  & 74.78  & -20.65  & 42.39  & -17.74  & 29.56  & -15.24  & 69.12  & -11.56  & 66.31  & -10.83  & 67.55  & -11.87  \\
    & 100 & 97.90  & -1.51  & 94.04  & -3.27  & 78.72  & -15.54  & 43.17  & -16.65  & 29.86  & -15.15  & 77.21  & -4.20  & 74.03  & -8.87  & 70.70  & -9.31  \\
    & 200  & 98.43  & -1.01  & 92.64  & -5.20  & 84.71  & -7.90  & 45.91  & -13.90  & 30.80  & -12.24  & 78.31  & -1.64  & 69.36  & -13.13  & 71.45  & -7.86  \\
    & 300 & 98.50  & 0.00  & 95.07  & -2.48  & 85.37  & -7.53  & 46.28  & -12.25  & 33.66  & -10.56  & 80.32  & 2.24  & 73.87  & -4.90  & 73.29  & -5.07  \\
    \hline
    \end{tabular}
    }
  \caption{Comparison of the TIL performance across different memory sizes.}
  \label{tab:sup_memory_til}
\end{table*}

\end{document}